\pdfoutput=1
\documentclass[11pt]{article}
\usepackage{acl}
\usepackage{times}
\usepackage{makecell}
\usepackage{latexsym}
\usepackage{paralist}
\usepackage{amsmath}
\usepackage{multirow}
\usepackage{hyperref}
\usepackage{booktabs}
\usepackage{graphicx}
\usepackage{amssymb}
\usepackage{verbatim}
\usepackage{amsmath}
\usepackage{adforn}
\usepackage[T1]{fontenc}
\usepackage{bm}
\usepackage[ruled,vlined]{algorithm2e}
\usepackage[utf8]{inputenc}
\usepackage{colortbl}
\usepackage{microtype}
\usepackage{inconsolata}
\definecolor{Ocean}{RGB}{129,194,234}

\definecolor{deepcarmine}{rgb}{0.66, 0.13, 0.24}
\definecolor{yellow-green}{rgb}{0.6, 0.8, 0.2}

\title{Joyful: Joint Modality Fusion and Graph Contrastive Learning for Multimodal Emotion Recognition}

\author{Dongyuan Li, Yusong Wang, Kotaro Funakoshi, and Manabu Okumura \\
        Tokyo Institute of Technology, Tokyo, Japan\\
        \{lidy, wangyi, funakoshi, oku\}@lr.pi.titech.ac.jp}
\begin{document}
\maketitle
\begin{abstract}
Multimodal emotion recognition aims to recognize emotions for each utterance of multiple modalities, which has received increasing attention for its application in human-machine interaction. 
Current graph-based methods fail to simultaneously
depict global contextual features and local diverse uni-modal features in a dialogue. 
Furthermore, with the number of graph layers increasing, they easily fall into over-smoothing.
In this paper, we propose a method for \underline{\textbf{jo}}int modalit\underline{\textbf{y}} \underline{\textbf{fu}}sion and graph contrastive \underline{\textbf{l}}earning for
multimodal emotion recognition (\textsc{Joyful}), where multimodality fusion, contrastive learning, and emotion recognition are jointly optimized.
Specifically, we first design a new multimodal fusion mechanism that can provide deep interaction and fusion between the global contextual and uni-modal specific features.
Then, we introduce a graph contrastive learning framework with inter-view and intra-view contrastive losses to learn more distinguishable representations for samples with different sentiments. 
Extensive experiments on three benchmark datasets indicate that \textsc{Joyful} achieved state-of-the-art (SOTA) performance compared to all baselines.
\footnote{Code is released on Github (\href{https://anonymous.4open.science/r/MERC-7F88}{https://anonymous/MERC}).}
\end{abstract}

\section{Introduction}

\textit{``Integration of information from multiple sensory channels is crucial for understanding tendencies and reactions in humans''}~\citep{Science}.
Multimodal emotion recognition in conversations (MERC) aims exactly to identify and track the emotional state of each utterance from heterogeneous visual, audio, and text channels.
Due to its potential applications in creating human-computer interaction systems~\citep{Li2022},
social media analysis~\citep{GuptaMMMMBMM22,EMP}, and recommendation systems~\citep{SinghDS022}, MERC has received increasing attention in the natural language processing (NLP) community~\citep{Poria1,Poria2}, which even has the potential to be widely applied in
other tasks such as question answering~\cite{DBLP:conf/acl/OssowskiH23,DBLP:conf/lrec/WangSZY22,DBLP:journals/corr/abs-2206-15030}, text generation~\cite{DBLP:conf/acl/LiangMXW0023,DBLP:conf/acl/ZhangKO23,DBLP:conf/coling/LiYFO22} and bioinformatics~\cite{DBLP:conf/bionlp/NicolsonDK23,DBLP:conf/coling/YouLOS22}.

\begin{figure}[!t]
\includegraphics[width=0.46\textwidth]{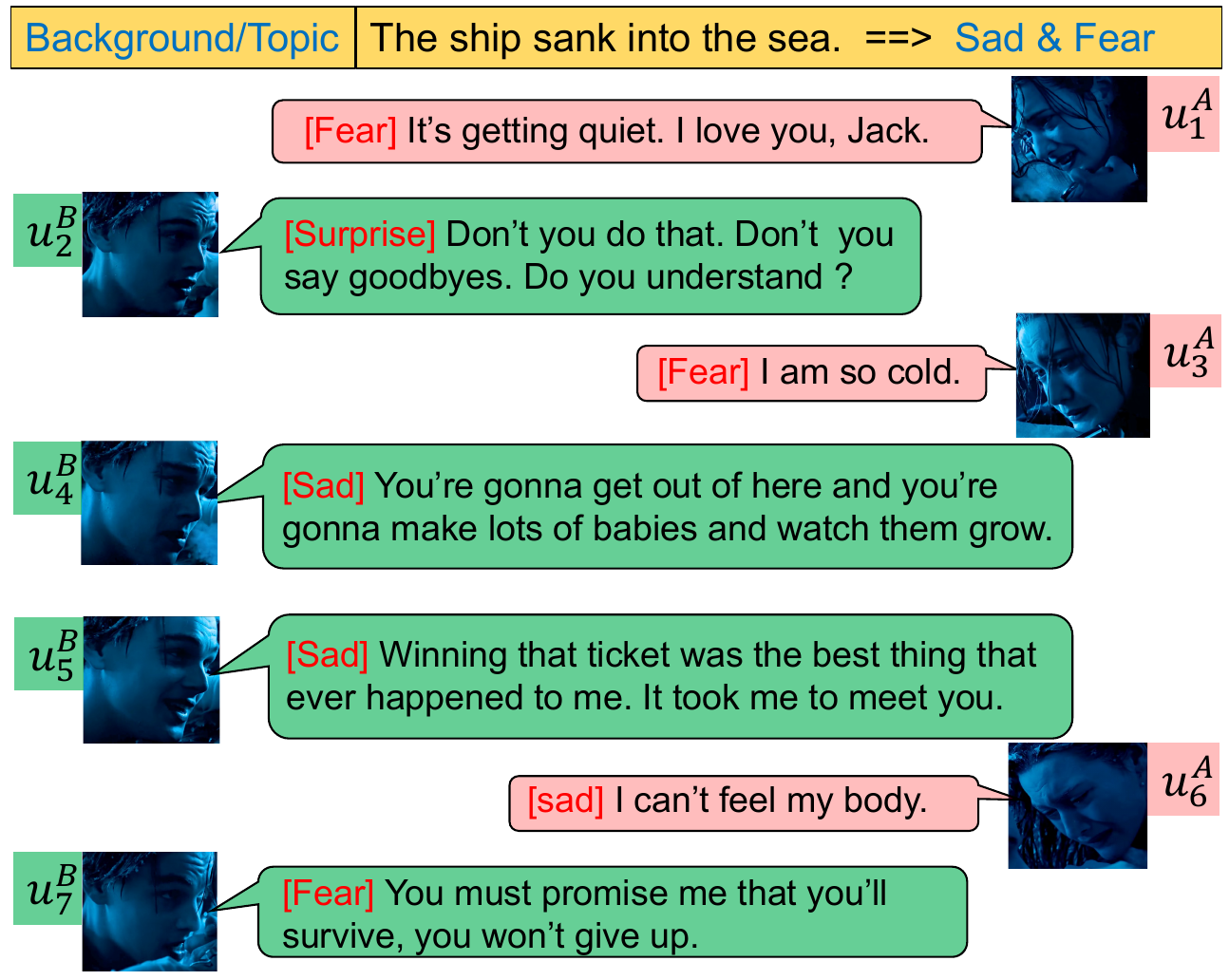}
\centering
\caption{Emotions are affected by multiple uni-modal,  global contextual, intra- and inter-person dependencies. Images are from the movie ``Titanic''. }
\label{example}
\end{figure}

Figure~\ref{example} shows that emotions expressed in a dialogue are affected by three main factors:
1) multiple uni-modalities (different modalities complete each other to provide a more informative utterance representation);
2) global contextual information ($u_{3}^{A}$ depends on the topic ``The ship sank into the sea'', indicating fear); and 
3) intra-person and inter-person dependencies ($u_{6}^{A}$ becomes sad affected by sadness in $u_{4}^{B}\& u_{5}^{B}$).
Depending on how to model intra-person and inter-person dependencies, 
current MERC methods can be categorized into Sequence-based and Graph-based methods.
The former~\citep{FE2E, maoetal, liang-etal} use recurrent neural networks or Transformers to model the temporal interaction between utterances.
However, they failed to distinguish intra-speaker and inter-speaker dependencies and easily lost uni-modal specific features by the cross-modal attention mechanism~\citep{CAttention}.
Graph structure~\citep{joshi-etal, MMGCN} solves these issues by using edges between nodes (speakers) to distinguish intra-speaker and inter-speaker dependencies.
Graph Neural Networks (GNNs) further help nodes learn common features by aggregating information from neighbours while maintaining their uni-modal specific features.

Although graph-based MERC methods have achieved great success, there still remain problems that need to be solved:
1) Current methods directly aggregate features of multiple modalities~\citep{joshi-etal} or project modalities into a latent space to learn representations~\citep{li2022clmlfa}, 
which ignores the diversity of each modality and fails to capture richer semantic information from each modality.
They also ignore global contextual information during the feature fusion process, leading to poor performance.
2) Since all graph-based methods adopt GNN~\citep{ScarselliGTHM09} or Graph Convolutional Networks (GCNs)~\citep{KipfW17}, with the number of layers deepening, the phenomenon of over-smoothing starts to appear, resulting in the representation of similar sentiments being indistinguishable.
3) Most methods use a two-phase pipeline~\citep{FuOWGSLD21,joshi-etal}, 
where they first extract and fuse uni-modal features as utterance representations and then fix them as input for graph models.
However, the two-phase pipeline will lead to sub-optimal performance since the fused representations are fixed and cannot be further improved to benefit from the downstream supervisory signals.

To solve the above-mentioned problems, we propose \textbf{Jo}int multimodalit\textbf{y} \textbf{fu}sion and graph contrastive \textbf{l}earning for MERC (\textsc{Joyful}), where multimodality fusion, graph contrastive learning (GCL), and multimodal emotion recognition are jointly optimized in an overall objective function. 
1) We first design a new multimodal fusion mechanism that can simultaneously learn and fuse a global contextual representation and uni-modal specific representations.
For the global contextual representation,
we smooth it with a proposed topic-related vector to maintain its consistency, where the topic-related vector is temporally updated since the topic usually changes.
For uni-modal specific representations, we project them into a shared subspace to fully explore their richer semantics without losing alignment with other modalities.
2) To alleviate the over-smoothing issue of deeper GNN layers, inspired by \citet{YouCSCWS20}, that showed contrastive learning could provide more distinguishable node representations to benefit various downstream tasks, 
we propose a cross-view GCL-based framework to alleviate the difficulty of categorizing similar emotions, which helps to learn more distinctive utterance representations by making samples with the same sentiment cohesive and those with different sentiments mutually exclusive.
Furthermore, graph augmentation strategies are designed to improve \textsc{Joyful}'s robustness and generalizability.
3) We jointly optimize each part of \textsc{Joyful} in an \textit{end-to-end} manner to ensure global optimized performance.
The main contributions of this study can be summarized as follows:

\textbullet\, We propose a novel joint leaning framework for MERC, where multimodality fusion, GCL, and emotion recognition are jointly optimized for global optimal performance. 
Our new multimodal fusion mechanism can obtain better representations by simultaneously depicting global contextual and local uni-modal specific features.

\textbullet\, To the best of our knowledge, \textsc{Joyful} is the first method to utilize graph contrastive learning for MERC, which significantly improves the model's ability to distinguish different sentiments. 
Multiple graph augmentation strategies further improve the model's stability and generalization.
     
\textbullet\, Extensive experiments conducted on three multimodal benchmark datasets demonstrated the effectiveness and robustness of \textsc{Joyful}.

\section{Related Work}
\subsection{Multimodal Emotion Recognition}

Depending on how to model the context of utterances, existing MERC methods are categorized into three classes:
Recurrent-based methods~\citep{MajumderPHMGC19,maoetal} adopt RNN or LSTM to model the sequential context for each utterance.
Transformers-based methods~\citep{ling,liang-etal,le-etal} use Transformers with cross-modal attention to model the intra- and inter-speaker dependencies. 
Graph-based methods~\cite{joshi-etal,ZhangJZLLZZ21,FuOWGSLD21} can control context information for each utterance and provide accurate intra- and inter-speaker dependencies, achieving SOTA performance on many MERC benchmark datasets.

\begin{figure*}[!ht]
\centering
\includegraphics[width=1\textwidth]{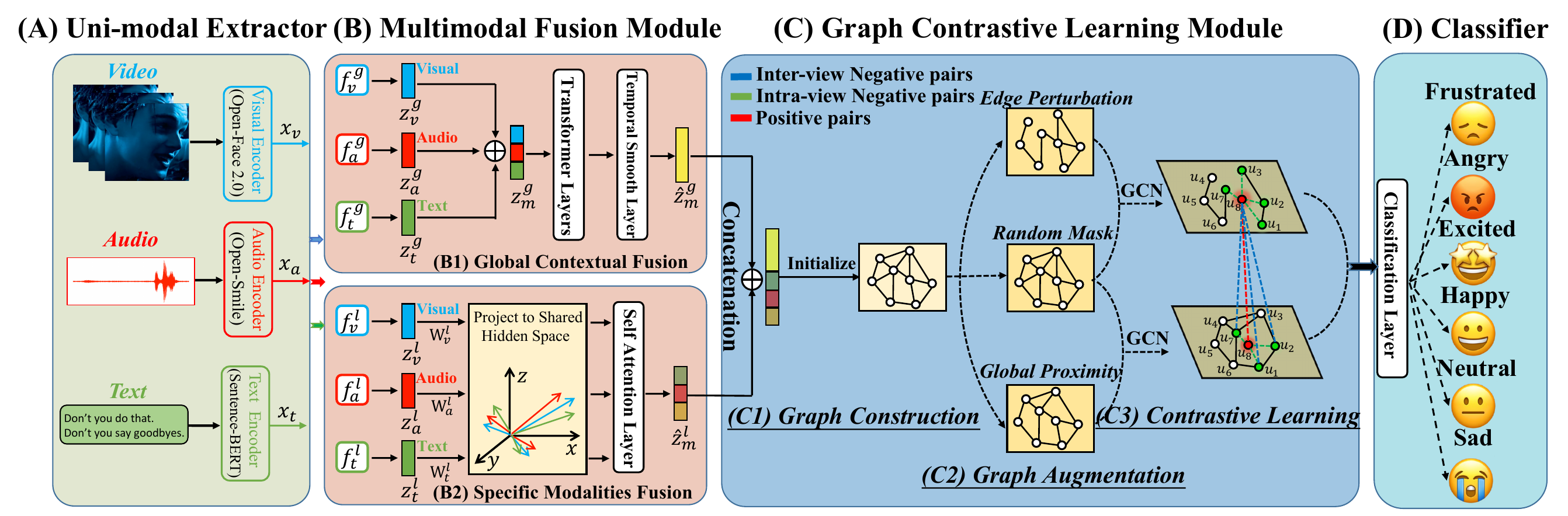}
\caption{Overview of \textsc{Joyful}. We first extract uni-modal features, fuse them using a multimodal fusion module, and use them as input of the GCL-based framework to learn better representations for emotion recognition.}
\label{overview}
\end{figure*}

\subsection{Multimodal Fusion Mechanism}
Learning effective fusion mechanisms is one of the core challenges in multimodal learning~\citep{shankar-2022-multimodal}. 
By capturing the interactions between different modalities more reasonably, deep models can acquire more comprehensive information. 
Current fusion methods can be classified into aggregation-based~\citep{WuZZWX21,GuoKZWW21}, alignment-based~\citep{LiuSYZX20,li2022clmlfa}, and their mixture~\citep{MMGCN,NagraniYAJSS21}.
Aggregation-based fusion methods~\citep{ZadehCPCM17,ChenSSW20} adopt concatenation, tensor fusion and memory fusion to combine multiple modalities. 
Alignment-based fusion centers on latent cross-modal adaptation, which adapts streams from one modality to another~\citep{cvpr022}.
Different from the above methods, we learn global contextual information by concatenation while fully exploring the specific patterns of each modality in an alignment manner.

\subsection{Graph Contrastive Learning}

GCL aims to learn representations by maximizing feature consistency under differently augmented views, that exploit data- or task-specific augmentations, to inject the desired feature invariance~\cite{YouCSCWS20}.
GCL has been well used in the NLP community via self-supervised and supervised settings.
Self-supervised GCL first creates augmented graphs by edge/node deletion and insertion~\citep{ZengX21}, or attribute masking~\citep{ZhangZSKK22}. 
It then captures the intrinsic patterns and properties in the augmented graphs without using human provided labels. 
Supervised GCL designs adversarial~\citep{SunQDLZ22} or geometric~\citep{LiZ0D022} contrastive loss to make full use of label information.
For example, \citet{DBLP:conf/aaai/LiYQ22} first used supervised CL for emotion recognition, greatly improving the performance.
Inspired by previous studies, we jointly consider self-supervised (suitable graph augmentation) and supervised (cross-entropy) manners to fully explore graph structural information and downstream supervisory signals.

%_________________________________________________%
%             Methodology
%_________________________________________________%
\section{Methodology}

Figure~\ref{overview} shows an overview of \textsc{Joyful}, which mainly consists of four components: 
(A) a \textit{uni-modal extractor}, 
(B) a \textit{multimodal fusion} (MF) module, 
(C) a \textit{graph contrastive learning} module, 
and (D) a \textit{classifier}.
Hereafter, we give formal notations and the task definition of \textsc{Joyful}, and introduce each component subsequently in detail.

\subsection{Notations and Task Definition}
In dialogue emotion recognition,
a training dataset 
$\mathcal{D} = 
\{(\mathcal{C}_{i},\mathcal{Y}_{i})\}_{i=1}^{N}$ is given, where $\mathcal{C}_{i}$ represents the $i$-th conversation, each conversation contains several utterances $\mathcal{C}_{i}=\{\bm{u}_{1},\dots,\bm{u}_{m}\}$, and $\mathcal{Y}_{i} \in \mathbf{Y}^m$, given label set $\mathbf{Y} = \{y_{1},\dots,y_{k}\}$ of $k$ emotion classes. 
Let $\mathbf{X}^{v}$, $\mathbf{X}^{a}$, $\mathbf{X}^{t}$ be the visual, audio, and text feature spaces, respectively.
The goal of MERC is to learn a function $\mathbf{F}:\mathbf{X}^{v} \times \mathbf{X}^{a} \times \mathbf{X}^{t} \to \mathbf{Y}$ that can recognize the emotion label for each utterance. 
% To make it easily understand, we list dimensions of mathematical symbols in Appendix~\ref{Dimension}.
We utilize three widely used multimodal conversational benchmark datasets, namely IEMOCAP, MOSEI, and MELD, to evaluate the performance of our model. Please see Section~\ref{dataset_description} for their detailed statistical information. 

\subsection{Uni-modal Extractor}
\label{3.2}

For IEMOCAP~\citep{BussoBLKMKCLN08}, video features $\bm{x}_{v}$ $\in$ $\mathbb{R}^{512}$, audio features $\bm{x}_{a} \in \mathbb{R}^{100}$, and text features $\bm{x}_{t} \in \mathbb{R}^{768}$ are obtained from OpenFace~\citep{BaltrusaitisZLM18}, OpenSmile~\citep{1874246} and SBERT~\citep{2019-sentence}, respectively.
For MELD~\citep{PoriaHMNCM19}, $\bm{x}_{v} \in \mathbb{R}^{342}$, $\bm{x}_{a} \in \mathbb{R}^{300}$, and $\bm{x}_{t} \in \mathbb{R}^{768}$ are obtained from DenseNet~\citep{DBLP:conf/cvpr/HuangLMW17}, OpenSmile, and TextCNN~\citep{DBLP:conf/emnlp/Kim14}.
For MOSEI~\citep{MorencyCPLZ18},  $\bm{x}_{v} \in \mathbb{R}^{35}$,  $\bm{x}_{a} \in \mathbb{R}^{80}$, and $\bm{x}_{t} \in \mathbb{R}^{768}$ are obtained from TBJE~\citep{2020-transformer}, LibROSA~\citep{RaguramanRV19}, and SBERT. 
Textual features are sentence-level static features. 
Audio and visual modalities are utterance-level features by averaging all the token features.

\subsection{Multimodal Fusion Module}

Though the uni-modal extractors can capture long-term temporal context, 
they are unable to handle feature redundancy and noise due to the modality gap.
Thus, we design a new multimodal fusion module (Figure~\ref{overview} (B)) to inherently separate multiple modalities into two disjoint parts, contextual representations and specific representations, to extract the consistency and specificity of heterogeneous modalities collaboratively and individually. 

\subsubsection{Contextual Representation Learning}

Contextual representation learning aims to explore and learn hidden contextual intent/topic knowledge of the dialogue, which can greatly improve the performance of \textsc{Joyful}.
In Figure~\ref{overview} (B1), we first project all uni-modal inputs $\bm{x}_{\{v,a,t\}}$ into a latent space by using three separate connected deep neural networks $f_{\{v,a,t\}}^{g}(\cdot)$ to obtain hidden representations $\bm{z}_{\{v,a,t\}}^{g}$.
Then, we concatenate them as $\bm{z}_{m}^{g}$ and apply it to a multi-layer transformer
to maximize the correlation between multimodal features, where we learn a global contextual multimodal representation $\hat{\bm{z}}_{m}^{g}$.
Considering that the contextual information will change over time,  we design a temporal smoothing strategy for $\hat{\bm{z}}_{m}^{g}$ as
\begin{equation}\label{eq1} 
        \mathcal{J}_{smooth} = \|\hat{\bm{z}}_{m}^{g}- \bm{z}_{con} \|^{2}, 
\end{equation}
where $\bm{z}_{con}$ is the topic-related vector describing the high-level global contextual information without requiring topic-related inputs, following the definition in \citet{joshi-etal}.
We update the ($i$+1)-th utterance as  $\bm{z}_{con} \leftarrow \bm{z}_{con} + e^{\eta*i} \hat{\bm{z}}_{m}^{g}$, and $\eta$ is the exponential smoothing parameter~\citep{shazeer2018adafactor}, indicating that more recent information will be more important.

To ensure fused contextual representations capture enough details from hidden layers, \citet{MISA_paper} minimized the reconstruction error between fused representations with hidden representations. 
Inspired by their work, to ensure that $\hat{\bm{z}}_{m}^{g}$ contains essential modality cues for downstream emotion recognition, we reconstruct $\bm{z}_{m}^{g}$ from $\hat{\bm{z}}_{m}^{g}$ by minimizing their Euclidean distance:
\begin{equation}
    \mathcal{J}_{rec}^{g} = \| \hat{\bm{z}}_{m}^{g} - \bm{z}_{m}^{g}\|^{2}. \label{eq2}
\end{equation}

\subsubsection{Specific Representation Learning}

Specific representation learning aims to fully explore specific information from each modality to complement one another.
Figure~\ref{overview} (B2) shows that we first use three fully connected deep neural networks $f_{\{v,a,t\}}^{\ell}(\cdot)$ to project uni-modal embeddings $\bm{x}_{\{v,a,t\}}$ into a hidden space with representations as $\bm{z}_{\{v,a,t\}}^{\ell}$. 
Considering that visual, audio, and text features are extracted with different encoding methods, directly applying multiple specific features as an input for the downstream emotion recognition task will degrade the model's accuracy.
To solve it, 
the multimodal features are projected into a shared subspace, and a shared trainable basis matrix is designed to learn aligned representations for them.
Therefore, the multimodal features can be fully integrated and interacted to mitigate feature discontinuity and remove noise across modalities.
We define a shared trainable basis matrix $\mathbf{B}$ with $q$ basis vectors as 
$\mathbf{B}=(\bm{b}_{1}, \dots, \bm{b}_{
q})^T \in \mathbb{R}^{q \times d_{b}}$
with $d_{b}$ representing the dimensionality of each basis vector.
Here, $T$ indicates transposition.
Then, $\bm{z}_{\{v,a,t\}}^{\ell}$ and $\mathbf{B}$ are projected into the shared subspace: 
\begin{equation}
\tilde{\bm{z}}_{\{v,a,t\}}^{\ell}=\mathbf{W}_{\{v,a,t\}} \bm{z}_{\{v,a,t\}}^{\ell},\,\,\,\,\,  \widetilde{\mathbf{B}}=\mathbf{B} \mathbf{W}_{b},
\end{equation}
where $\mathbf{W}_{\{v,a,t,b\}} $ are trainable parameters.
To learn new representations for each modality, we calculate the cosine similarity between them and $\mathbf{B}$ as
\begin{equation}
S_{ij}^{\{v,a,t\}}=(\tilde{\bm{z}}_{\{v,a,t\}}^{\ell})_{i} \cdot \widetilde{\bm{b}}_{j},
\end{equation}
where 
$S_{ij}^{v}$ denotes the similarity between the $i$-th visual feature $(\tilde{\bm{z}}_{v}^{\ell})_{i}$ and the $j$-th basis vector representation $\widetilde{\bm{b}}_{j}$.
To prevent inaccurate representation learning caused by an excessive weight of a certain item, the similarities are further normalized by
\begin{equation}
    S_{ij}^{\{v,a,t\}}=\frac{\exp{(S_{ij}^{\{v,a,t\}})}}{\sum_{k=1}^{q} \exp{(S_{ik}^{\{v,a,t\}})}}. 
\end{equation}

Then, the new representations are obtained as
\begin{equation}
    (\hat{\bm{z}}_{\{v,a,t\}}^{\ell})_{i}=\sum_{k=1}^{q} S_{ik}^{\{v,a,t\}} \cdot \widetilde{\bm{b}}_{k},
\end{equation}
where $\hat{\bm{z}}_{\{v,a,t\}}^{\ell}$ are new representations, and we also use reconstruction loss for their combinations
\begin{equation}
    \mathcal{J}_{rec}^{\ell} = \| \hat{\bm{z}}_{m}^{\ell} - \bm{z}_{m}^{\ell}\|^{2}, \label{eq3}
\end{equation}
where \textit{Concat( , )} indicating the concatenation, i.e., $\hat{\bm{z}}_{m}^{\ell}$=$\textit{Concat}(\hat{\bm{z}}_{v}^{\ell},\hat{\bm{z}}_{a}^{\ell},\hat{\bm{z}}_{t}^{\ell})$, 
$\bm{z}_{m}^{\ell}$ = $\textit{Concat}(\bm{z}_{v}^{\ell},\bm{z}_{a}^{\ell}, \bm{z}_{t}^{\ell})$.

Finally, we define the multimodal fusion loss by combining Eqs.(\ref{eq1}), (\ref{eq2}), and (\ref{eq3}) as:
\begin{equation}
    \mathcal{L}_{mf} = \mathcal{J}_{smooth} + \mathcal{J}_{rec}^{g} + \mathcal{J}_{rec}^{\ell}. \label{cm1}
\end{equation}

\subsection{Graph Contrastive Learning Module}

\begin{figure}[!ht]
\centering
\includegraphics[width=0.47\textwidth]{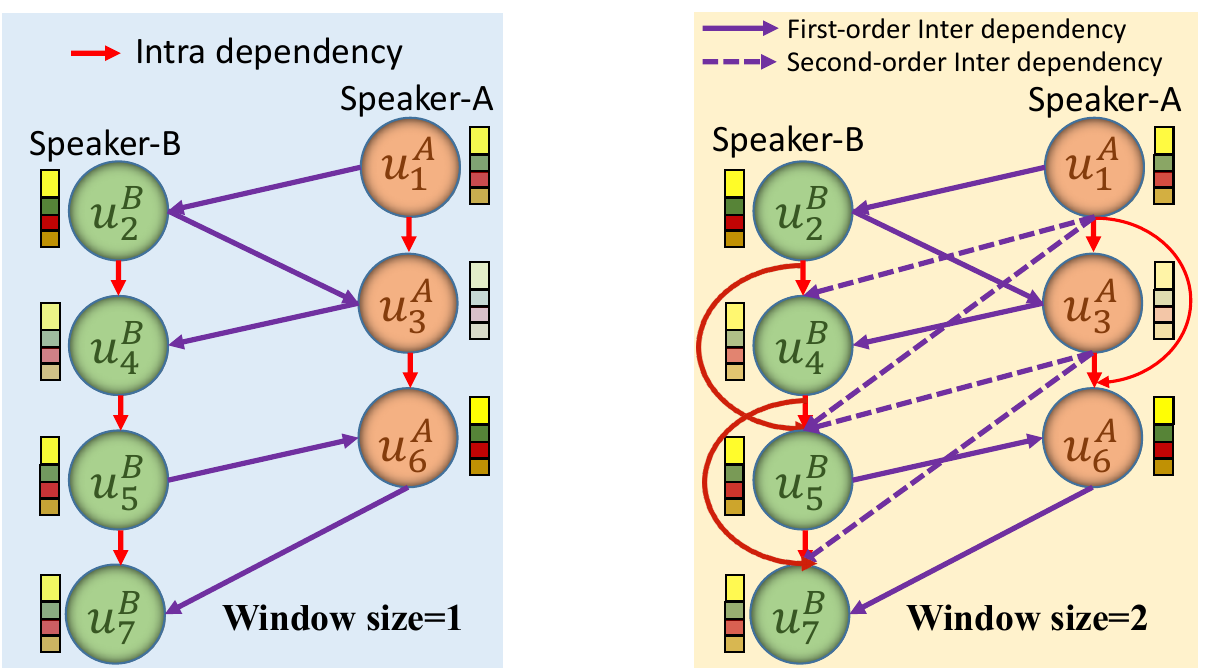}
\caption{An example of graph construction.}
\label{node_example}
\end{figure}

%\noindent 
\subsubsection{Graph Construction} 

Graph construction aims to establish relations between past and future utterances that preserve both intra- and inter-speaker dependencies in a dialogue. 
We define the $i$-th dialogue with $P$ speakers as $\mathcal{C}_{i}$ = $\{\mathcal{U}^{S_{1}}, \dots, \mathcal{U}^{S_{P}}\}$, where $\mathcal{U}^{S_{i}}$ = $\{\bm{u}_{1}^{S_{i}},\dots,\bm{u}_{m}^{S_{i}}\}$ represents the set of utterances spoken by speaker $S_{i}$.
Following \citet{dialoguegcn}, we define a graph with nodes representing utterances and directed edges representing their relations:  $\mathcal{R}_{ij}= \bm{u}_{i} \to \bm{u}_{j}$, where the arrow represents the speaking order. 
\textit{Intra-Dependency} ($\mathcal{R}_{intra}\in \{\mathcal{U}^{S_{i}} \to \mathcal{U}^{S_{i}}\}$) represents intra-relations between the utterances (red lines), 
and \textit{Inter-Dependency} ($\mathcal{R}_{inter} \in \{\mathcal{U}^{S_{i}} \to \mathcal{U}^{S_{j}}\}, i \neq j$) represents the inter-relations between the utterances (purple lines), as shown in Figure~\ref{node_example}.
All nodes are initialized by concatenating contextual and specific representations as $\bm{h}_{m}$ = \textit{Concat}($\hat{\bm{z}}_{m}^{g},\hat{\bm{z}}_{m}^{\ell}$). 
And we show that window size is a hyper-parameter that controls the context information for each utterance and provide accurate intra- and inter-speaker dependencies.

\subsubsection{Graph Augmentation}

Graph Augmentation (GA): 
Inspired by \citet{DGCRL}, creating two augmented views by using different ways to corrupt the original graph can provide highly heterogeneous contexts for nodes.
% The motivation for using GA is that 
% creating two augmented views by corrupting the original graph in different ways can provide highly heterogeneous contexts for nodes~\citep{DGCRL}. 
By maximizing the mutual information between two augmented views, we can improve the robustness of the model and obtain distinguishable node representations~\citep{YouCSCWS20}.
However, there are no universally appropriate GA methods for various downstream tasks~\citep{InfoGCL}, which motivates us to
design specific GA strategies for MERC.
Considering that MERC is sensitive to initialized representations of utterances, intra-speaker and inter-speaker dependencies, we design three corresponding GA methods:
\begin{itemize}
\item[-] \textbf{Feature Masking} (FM):
given the initialized representations of utterances, we randomly select $p$ dimensions of the initialized representations and mask their elements with zero, which is expected to enhance the robustness of \textsc{Joyful} to multimodal feature variations;
\item[-] \textbf{Edge Perturbation} (EP):
given the graph $\mathcal{G}$, we randomly drop and add $p\%$ of intra- and inter-speaker edges, which is expected to enhance the robustness of \textsc{Joyful} to local structural variations;
\item[-] \textbf{Global Proximity} (GP):
given the graph $\mathcal{G}$, we first use the Katz index~\citep{katz1953new} to calculate high-order similarity between 
intra- and inter-speakers, and randomly add $p\%$ high-order edges between 
speakers, which is expected to enhance the robustness of \textsc{Joyful} to global structural variations (Examples in Appendix~\ref{High-order}).
\end{itemize}

We propose a hybrid scheme for generating graph views on both structure and attribute levels to provide diverse node contexts for the contrastive objective. 
Figure~\ref{overview} (C) 
shows that the combination of (FM $\&$ EP) and (FM $\&$ GP)
are adopted to obtain two correlated views.

\subsubsection{Graph Contrastive Learning}

Graph contrastive learning adopts an $L$-th layer GCNs as a graph encoder to extract node hidden representations $\mathbf{H}^{(1)}$ = \{$ \bm{h}^{(1)}_{1}$, $\dots$, $\bm{h}^{(1)}_{m}$\} 
and $\mathbf{H}^{(2)}$ = \{$\bm{h}^{(2)}_{1}$, $\dots$, $\bm{h}^{(2)}_{m}$\} for two augmented graphs, where $\bm{h}_{i}$ is the hidden representation for the $i$-th node. 
We follow an iterative neighborhood aggregation (or message passing) scheme to capture the structural information within the nodes' neighborhood. 
Formally, the propagation and aggregation of the $\ell$-th GCN layer is:
\begin{align}
\bm{a}_{(i,\,\ell)} =& \,\,\,\textrm{AGG}_{(\ell)\,}(\{\bm{h}_{(j,\,\ell-1)}|j \in \mathbf{N}_{i}\}) \\
\bm{h}_{(i,\,\ell)} =& \,\,\,\textrm{COM}_{(\ell)\,}(\bm{h}_{(i,\,\ell-1)} \oplus \bm{a}_{(i,\,\ell)}) \label{Eq.10}, 
\end{align}
where $\bm{h}_{(i,\,\ell)}$ is the embedding of the $i$-th node at the $\ell$-th layer, $\bm{h}_{(i,\,0)}$ is the initialization of the $i$-th utterance, $\mathbf{N}_i$ represents all neighbour nodes of the $i$-th node, and $\text{AGG}_{(\ell)}(\cdot)$ and $\text{COM}_{(\ell)}(\cdot)$ are aggregation and combination of the $\ell$-th GCN layer~\citep{HamiltonYL17}.
For convenience, we define $\boldsymbol{h}_{i} = \boldsymbol{h}_{(i,L)}$. 
After the $L$-th GCN layer, final node representations of two views are $\mathbf{H}^{(1)}$ / $\mathbf{H}^{(2)}$.

In Figure~\ref{overview} (C3), we design the intra- and inter-view graph contrastive losses to learn distinctive node representations. 
We start with the inter-view contrastiveness, which pulls closer the representations of the same nodes in two augmented views while pushing other nodes away,
as depicted by the red and blue dash lines in Figure~\ref{overview} (C3).
Given the definition of our positive and negative pairs as 
$(\bm{h}_{i}^{(1)},\bm{h}_{i}^{(2)})^{+}$ and $(\bm{h}_{i}^{(1)},\bm{h}_{j}^{(2)})^{-}$, where $i \neq j$, 
the inter-view loss for the $i$-th node is formulated as:
\begin{equation}\label{inter}
\begin{aligned}
    \mathcal{L}_{inter}^{i} = 
      -\log \frac{\exp(\text{sim}(\bm{h}_{i}^{(1)}, \bm{h}_{i}^{(2)}))}
      {\sum\limits_{j=1}^m \exp(\text{sim}(\bm{h}_{i}^{(1)}, \bm{h}_{j}^{(2)}))},
\end{aligned}
\end{equation}
where $\text{sim}(\cdot,\cdot)$ denotes the similarity between two vectors, i.e., the cosine similarity in this paper.

Intra-view contrastiveness regards all nodes except the anchor node as negatives within a particular view
(green dash lines in Figure~\ref{overview} (C3)),
as defined $(\bm{h}_{i}^{(1)},\bm{h}_{j}^{(1)})^{-}$ where $i \neq j$. 
The intra-view contrastive loss for the $i$-th node is defined as:
\begin{equation}\label{intra}
\begin{aligned}
    \mathcal{L}_{intra}^{i} = -\log 
      \frac{\exp(\text{sim}(\bm{h}_{i}^{(1)}, \bm{h}_{i}^{(2)}))}
      {\sum\limits_{j=1}^m \exp(\text{sim}(\bm{h}_{i}^{(1)}, \bm{h}_{j}^{(1)}))}.
\end{aligned}
\end{equation}

By combining the inter- and intra-view contrastive losses of Eqs.(\ref{inter}) and (\ref{intra}), the contrastive objective function $\mathcal{L}_{ct}$ is formulated as:
\begin{equation}
\begin{aligned}
    \mathcal{L}_{ct} = 
    \frac{1}{2m}\sum\limits_{i=1}^m
    (\mathcal{L}_{inter}^{i} + \mathcal{L}_{intra}^{i}). \label{cm2}
\end{aligned}
\end{equation}

\subsection{Emotion Recognition Classifier}

We use cross-entropy loss for classification as:
\begin{equation}
    \mathcal{L}_{ce} = -\frac{1}{m} \sum_{i=1}^{m}\sum_{j=1}^{k} y^{j}_{i} \log{ (\hat{y}_{i}^{j})}, \label{cm3}
\end{equation}
where $k$ is the number of emotion classes,  $m$ is the number of utterances, $\hat{y}_{i}^{j}$ is the $i$-th predicted label, and $y_{i}^{j}$ is the $i$-th ground truth of $j$-th class. 

Above all, combining the MF loss of Eq.(\ref{cm1}), contrastive loss of Eq.(\ref{cm2}), and classification loss of  Eq.(\ref{cm3}) together, the final objective function is
\begin{equation}
    \mathcal{L}_{all} = \boldsymbol{\alpha} \mathcal{L}_{mf}  + \boldsymbol{\beta} \mathcal{L}_{ct}\label{cmall} + \mathcal{L}_{ce},  
\end{equation}
where $\boldsymbol{\alpha}$ and $\boldsymbol{\beta}$ are the trade-off hyper-parameters. 
We give our pseudo-code in Appendix~\ref{pseudo}.

\begin{table}[t]
\centering
\footnotesize
\setlength{\tabcolsep}{1mm}{ %Change weight of table
\begin{tabular}{lccc}
\toprule
\textit{\textbf{Dataset}}     
& \multicolumn{1}{c}{Train} & \multicolumn{1}{c}{\,\,\,\,Valid} & Test \\ 
\midrule 
\midrule
\multicolumn{1}{l}{IEMOCAP(4-way)}    & \multicolumn{1}{c}{3,200/108}  
& \multicolumn{1}{c}{400/12}   & 943/31 \\ 
\multicolumn{1}{l}{IEMOCAP(6-way)}    & \multicolumn{1}{c}{5,146/108}  
& \multicolumn{1}{c}{664/12}   & 1,623/31 \\ 
% \midrule
\multicolumn{1}{l}{MELD}  & \multicolumn{1}{c}{9,989/1,039} 
& \multicolumn{1}{c}{1,109/114}  & 2,80/2,610 \\ 
\multicolumn{1}{l}{MOSEI}  & \multicolumn{1}{c}{16,327/2,249} 
& \multicolumn{1}{c}{1,871/300}  & 4,662/679 \\ 
\bottomrule
\end{tabular}}
\caption{Utterances/Conversations of four datasets.} 
%IEMOCAP shows the number of utterances/conversations. }
\label{statistic}
\end{table}

\section{Experiments and Result Analysis}

\subsection{Experimental Settings}
\label{dataset_description}

\noindent \textbf{Datasets and Metrics.}
In Table~\ref{statistic}, 
IEMOCAP is a conversational dataset where each utterance was labeled with one of the six emotion categories (Anger, Excited, Sadness, Happiness, Frustrated and Neutral).
Following COGMEN, two IEMOCAP settings were used for testing, one with four emotions (Anger, Sadness, Happiness and Neutral) and one with all six emotions, where 4-way directly removes the additional two emotion labels (Excited and Frustrated). 
MOSEI was labeled with six emotion labels (Anger, Disgust, Fear, Happiness, Sadness, and Surprise).
For six emotion labels, we conducted two settings: \textit{binary classification} considers the target emotion as one class and all other emotions as another class,
and \textit{multi-label classification} tags multiple labels for each utterance.
MELD was labeled with six universal emotions (Joy, Sadness, Fear, Anger, Surprise, and Disgust).
We split the datasets into 70\%/10\%/20\% as training/validation/test data, respectively.
Following \citet{joshi-etal}, we used \textit{Accuracy} and \textit{Weighted F1-score} (WF1) as evaluation metrics.
Please note that the detailed label distribution of the datasets is given in Appendix~\ref{label_distribution}.

%The label distribution of all datasets are shown in Table~\ref{Dimensions-1}.

\noindent \textbf{Implementation Details.}
We selected the augmentation pairs (FM \& EP) and  (FM \& GP) for two views. 
We set the augmentation ratio $p$=20\% and smoothing parameter $\eta$=0.2, and applied the Adam~\citep{Adam} optimizer with an initial learning rate of 3$e$-5. 
For a fair comparison, we followed the default parameter settings of the baselines and repeated all experiments ten times to report the average accuracy.  
We conducted the significance by t-test with Benjamini-Hochberg~\citep{benjam} correction (Please see details in Appendix~\ref{B-H_analysis}).

\noindent \textbf{Baselines.}
Different MERC datasets have different best system results, 
following COGMEN, 
we selected SOTA baselines for each dataset. 
\underline{For IEMOCAP-4}, we selected Mult~\citep{Mult}, 
RAVEN~\citep{WangSLLZM19}, MTAG~\citep{YangWYZRZPM21}, PMR~\citep{LvCHDL21}, COGMEN and MICA~\citep{LiangLFZL21} as our baselines. 
\underline{For IEMOCAP-6}, we selected Mult, FE2E~\cite{FE2E}, DiaRNN~\cite{MajumderPHMGC19}, COSMIC~\citep{cosmic}, Af-CAN~\citep{WangHZZ21}, AGHMN~\citep{JiaoLK20}, COGMEN and RGAT~\citep{ishrelation} as our baselines. 
\underline{For MELD}, we selected DiaGCN~\citep{dialoguegcn}, DiaCRN~\citep{hu-etal-2021-dialoguecrn}, MMGCN~\citep{MMGCN}, UniMSE~\citep{hu-etal-2022-unimse}, COGMEN and MM-DFN~\citep{HuHWJM22} as baselines. 
\underline{For MOSEI}, we selected Mul-Net~\citep{abs-2002-08267}, TBJE~\citep{2020-transformer}, COGMEN and MR~\citep{TsaiMYSM20}.
%(Details in Appendix~\ref{Baselines}).

% \begin{figure}[!ht]
% \includegraphics[width=0.48\textwidth]{Figure/augmentation1.pdf}
% \centering
% \caption{WF1 gain when contrasting different augmentation pairs and no graph augmentation module.}
% \label{augmentation1}
% \end{figure}

\begin{figure*}[!ht]
\includegraphics[width=1\textwidth]{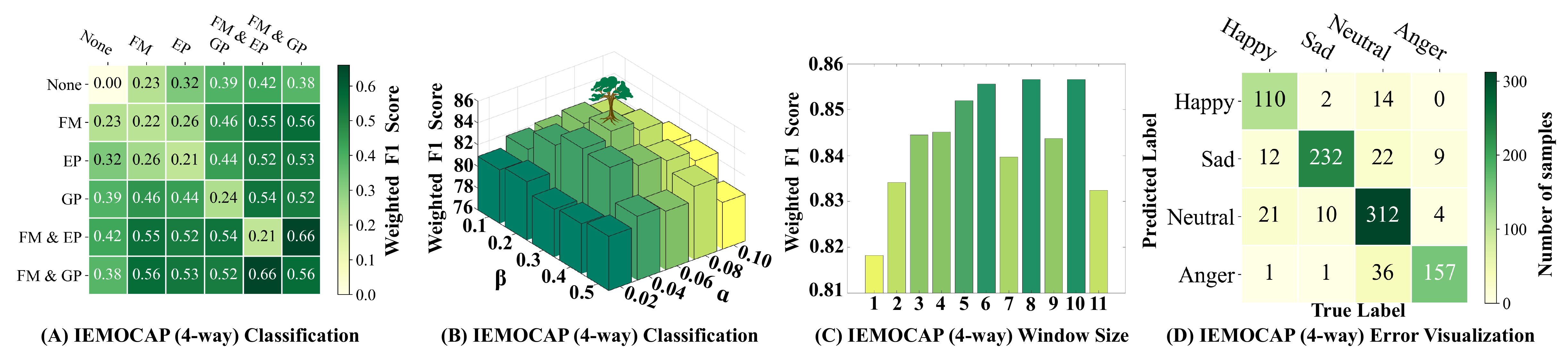}
\centering
\caption{(A) WF1 gain with different augmentation pairs; (B$\sim$C) Parameter tuning; (D) Imbalanced dataset.}
%and no graph augmentation module; (B$\sim$C) Parameter tuning; (D) Imbalanced dataset visualization.}
\label{augmentation1}
\end{figure*}

\subsection{Parameter Sensitive Study}
\label{Augmentation}
We first examined whether applying different data augmentation methods improves \textsc{Joyful}. 
% As shown in Figure~\ref{augmentation1}, 
We observed in Figure~\ref{augmentation1} (A) that 
1) %without any data augmentation, the GCL module will not improve the accuracy;
all data augmentation strategies are effective
2) applying augmentation pairs of the same type cannot result in the best performance; and 
3) applying augmentation pairs of different types improves performance.
Thus, we selected (FM $\&$ EP) and (FM $\&$ GP) as the default augmentation strategy since they achieved the best performance (More details please see Appendix~\ref{Data_augmentation}).

\textsc{Joyful} has three hyperparameters. $\boldsymbol{\alpha}$ and $\boldsymbol{\beta}$ determine the importance of MF and GCL in Eq.(\ref{cmall}), and window size controls the contextual length of conversations.
In Figure~\ref{augmentation1} (B), we observed how $\boldsymbol{\alpha}$ and $\boldsymbol{\beta}$ affect the performance of \textsc{Joyful} by varying $\boldsymbol{\alpha}$ from 0.02 to 0.10 in 0.02 intervals and $\boldsymbol{\beta}$ from 0.1 to 0.5 in 0.1 intervals.
The results indicated that \textsc{Joyful} achieved the best performance when $\boldsymbol{\alpha} \in  \left[0.06,0.08 \right]$ and $\boldsymbol{\beta}$ = 0.3. 
Figure~\ref{augmentation1} (C) shows that when ${window\_size}$ = 8, \textsc{Joyful} achieved the best performance.
A small window size will miss much contextual information, and a longer one contains too much noise, we set it as 8 in experiments (Details in Appendix~\ref{parameters}).

\begin{table}[!ht]
\scriptsize
\centering
\setlength{\tabcolsep}{0.33mm}{ %Change weight of table
\begin{tabular}{lcccccccc}
\toprule
\multirow{2}{*}{\textit{\textbf{Method}}} & \multicolumn{6}{c}{\textit{\textbf{ IEMOCAP 6-way (F1)}} $\,\uparrow$}                 & \multicolumn{2}{c}{\textit{\textbf{Average}} $\,\uparrow$} \\ 
\cmidrule(lr){2-7}\cmidrule(l){8-9}
& \textbf{\textit{Hap.}}  & \textbf{\textit{Sad.}} & \textbf{\textit{Neu.}} & \textbf{\textit{Ang.}} & \textbf{\textit{Exc.}}  
& \textbf{\textit{Fru.}}  & \textbf{\textit{Acc.}}  & \textbf{\textit{WF1}}\\ 
\midrule
\midrule
Mult   & 48.23 & 76.54 & 52.38 & 60.04 & 54.71 & 57.51 & 58.04 & 58.10\\
FE2E   & 44.82 & 64.98 & 56.09 & 62.12 & 61.02 & 57.14 & 58.30 & 57.69\\
%MFN                     & 34.12         & 70.53          & 52.11          & 66.85          & 62.12          & 62.56           & 60.18         & 59.91         \\ 
%ICON                    & 32.82          & 74.46          & 60.63          & 68.26          & 68.41          & 66.23           & 64.07         & 63.51         \\ 
DiaRNN             & 32.88          & 78.08          & 59.11          & 63.38          & 73.66          & 59.41           & 63.34         & 62.85         \\ 
%IEIN                    & 53.17          & 77.19          & 61.31          & 61.45          & 69.23          & 60.92           & 64.01         & 64.37         \\ 
COSMIC & 53.23 & 78.43 & 62.08 & 65.87 & 69.60 & 61.39 &64.88  &65.38\\
Af-CAN                 & 37.01         & 72.13          & 60.72          & 67.34          & 66.51          & 66.13           & 64.62         & 63.74         \\ 
%BERT-Base               & 37.09          & 59.53          & 51.73          & 54.32          & 54.26          & 55.83           & 54.21         & 53.31         \\ 
AGHMN                   & 52.10          & 73.30          & 58.40          & 61.91          & 69.72          & 62.31           & 63.58         & 63.54         \\ 
RGAT                    & 51.62          & 77.32          & 65.42          & 63.01          & 67.95          & 61.23           & 65.55         & 65.22         \\ 
COGMEN                  & 51.91          & 81.72          &  \cellcolor{olive!20} \textbf{68.61} & 66.02          &  \cellcolor{olive!20}\textbf{75.31} & 58.23           & 68.26         & 67.63         \\ 
\midrule
\textbf{\textsc{Joyful}}                 & \cellcolor{olive!20}$\textbf{\,\,\,60.94}^{\dag}$ & \cellcolor{olive!20}$\textbf{\,\,\,84.42}^{\dag}$  & 68.24          & \cellcolor{olive!20}$\textbf{\,\,\,69.95}^{\dag}$  & 73.54          & \cellcolor{olive!20}$\textbf{\,\,\,67.55}^{\dag}$   & \cellcolor{olive!20}$\textbf{\,\,\,70.55}^{\dag}$          & \cellcolor{olive!20}$\textbf{\,\,\,71.03}^{\dag}$          \\ 
\bottomrule
\end{tabular}}
\caption{Overall performance comparison on IEMOCAP (6-way) in the multimodal (A+T+V) setting.
Symbol $\dag$ indicates that \textsc{Joyful} significantly surpassed all baselines using t-test with $p<0.005$.}
\label{6-way}
\end{table} 

\begin{table}[!ht]
\scriptsize
\centering
\setlength{\tabcolsep}{2.3mm}{ %Change weight of table
\begin{tabular}{lccccc}
\toprule
\multirow{1}{*}{\textit{\textbf{Method}}} 
& \textit{\textbf{Happy}}     & \textit{\textbf{Sadness}}       & \textit{\textbf{Neutral}}       & \textit{\textbf{Anger}}         & \textit{\textbf{WF1}}          \\ 
\midrule
\midrule
Mult                & 88.4         & 86.3          & 70.5          & 87.3          & 80.4                 \\ 
%MFN             & 84.0          & 82.1          & 69.2          & 83.7          & 79.8                 \\ 
RAVEN                    & 86.2          & 83.2          & 69.4          & 86.5          & 78.6                  \\ 
MTAG                  & 85.9       & 80.1          & 64.2          & 76.8          & 73.9                  \\ 
PMR               & \cellcolor{olive!20}\textbf{89.2}           & 87.1          & 71.3          & 87.3          & 81.0                  \\ 
MICA                   & 83.7          & 75.5          & 61.8          & 72.6          & 70.7              \\ 
COGMEN                & 78.8         & 86.8          & 84.6          & 88.0          & 84.9                   \\ 
\midrule
\textbf{\textsc{Joyful}} &80.1                 & \cellcolor{olive!20}$\textbf{\,\,\,\,88.1}^{\dag}$ & \cellcolor{olive!20}$\textbf{\,\,\,85.1}^{\dag}$           & \cellcolor{olive!20}$\textbf{\,\,\,\,88.1}^{\dag}$  & \cellcolor{olive!20}$\textbf{\,\,\,\,85.7}^{\dag}$  
\\ 
\bottomrule
\end{tabular}}
\caption{Overall performance comparison on IEMOCAP (4-way) in the multimodal (A+T+V) setting.}
\label{4-way}
\end{table} 

\subsection{Performance of \textsc{Joyful}}

Tables~\ref{6-way} \& \ref{4-way} show that \textsc{Joyful} outperformed all baselines in terms of accuracy and WF1, improving 5.0\% and 1.3\% in WF1 for 6-way and 4-way, respectively.
Graph-based methods, COGMEN and \textsc{Joyful}, outperform Transformers-based methods, Mult and FE2E. 
Transformers-based methods cannot distinguish intra- and inter-speaker dependencies, distracting their attention to important utterances. 
Furthermore, they use the cross-modal attention layer, which can  enhance common features among modalities while losing  uni-modal specific features~\citep{CAttention}.
\textsc{Joyful} outperforms other GNN-based methods since it explored features from both the contextual and specific levels, and used GCL to obtain more distinguishable features. 
However, \textsc{Joyful} cannot improve in Happy for 4-way and in Excited for 6-way since samples in IEMOCAP were insufficient for distinguishing these similar emotions (Happy is 1/3 of Neutral in  Fig.~\ref{augmentation1} (D)). 
Without labels' guidance to re-sample or re-weight the underrepresented samples, self-supervised GCL, utilized in \textsc{Joyful}, cannot ensure distinguishable representations for samples of minor classes by only exploring graph topological information and vertex attributes.

\begin{table}[!ht]
\scriptsize
\centering
\setlength{\tabcolsep}{0.75mm}{ %Change weight of table
\begin{tabular}{lccccccc}
\toprule
\multirow{2}{*}{\textbf{Methods}} & \multicolumn{5}{c}{\textbf{Emotion Categories of MELD (F1) $\uparrow$}}              
& \multicolumn{2}{c}{\textbf{Average $\uparrow$}}
\\ 
\cmidrule(l){2-6} \cmidrule(l){7-8}
\multicolumn{1}{c}{}
& \textbf{\textit{Neu.}}  & \textbf{\textit{Sur.}}  & \textbf{\textit{Sad.}}  & \textbf{\textit{Joy}}  & \textbf{\textit{Anger}}  & \textbf{\textit{Acc.}}  & \textbf{\textit{WF1}}  \\ 
\midrule 
\midrule
DiaGCN &75.97 &46.05 &19.60 &51.20 &40.83 &58.62 &56.36  \\
DiaCRN &77.01 &50.10 &26.63 &52.77 &45.15 &61.11 &58.67\\
MMGCN & 76.33 & 48.15 & 26.74 & 53.02 & 46.09 &60.42 &58.31 \\
UniMSE & \underline{74.61} & \underline{48.21} & \underline{31.15} & \underline{54.04} & \underline{45.26} & \underline{59.39} & \underline{58.19}\\
COGMEN & \underline{75.31} & \underline{46.75} & \underline{33.52} & \underline{54.98} & \underline{45.81} & \underline{58.35} & \underline{58.66}  \\
MM-DFN &\cellcolor{olive!20}$\textbf{77.76}$ &50.69 & 22.93 & 54.78 &47.82 &62.49 & 59.46 \\
\midrule
\textbf{\textsc{Joyful}}& 76.80 &\cellcolor{olive!20}$\textbf{\,\,\,51.91}^{\dag}$  &\cellcolor{olive!20}$\textbf{\,\,\,41.78}^{\dag}$ &\cellcolor{olive!20}$\textbf{\,\,\,56.89}^{\dag}$ &\cellcolor{olive!20}$\textbf{\,\,\,50.71}^{\dag}$ &\cellcolor{olive!20}$\textbf{\,\,\,62.53}^{\dag}$ 
&\cellcolor{olive!20}$\textbf{\,\,\,61.77}^{\dag}$ \\ 
\bottomrule
\end{tabular}}
\caption{Results on MELD with the multimodal setting. 
\underline{Underline} indicates our reproduced results.} 
\label{performance_on_MELD}
\end{table}

\begin{table}[!ht]
\scriptsize
\centering
\setlength{\tabcolsep}{1.45mm}{ %Change weight of table
\begin{tabular}{lcccccc}
\toprule
\multirow{1}{*}{\textit{\textbf{Method}}} 
&\textbf{\textit{Happy}} &\textbf{\textit{Sadness}} &\textbf{\textit{Anger}} &\textbf{\textit{Fear}} &\textbf{\textit{Disgust}} &\textbf{\textit{Surprise}}\\
\midrule
\midrule

& \multicolumn{6}{c}{{\textbf{Binary Classification (F1)}}$\,\uparrow$} 
\\
\cmidrule(l){2-7}
Mul-Net            
& 67.9         & 65.5          & 67.2          & 87.6          & 74.7          & 86.0
\\
\multirow{1}{*}{TBJE}         
& 63.8         & 68.0          & 74.9          & 84.1          & 83.8          & 86.1 
\\
\multirow{1}{*}{MR}  & 65.9         & 66.7          & 71.0          & 85.9          & 80.4          & 85.9\\
\multirow{1}{*}{COGMEN}         
& 70.4         & 72.3          & 76.2          & 88.1          & 83.7          & 85.3 
\\
\multirow{1}{*}{\textsc{\textbf{Joyful}}}       
& \cellcolor{olive!20}$\textbf{\,\,\,\,71.7}^{\dag}$         & \cellcolor{olive!20} $\textbf{\,73.4}^{\dag}$         & \cellcolor{olive!20} $\textbf{\,78.9}^{\dag}$           & \cellcolor{olive!20}\textbf{88.2}           & \cellcolor{olive!20} $\textbf{\,85.1}^{\dag}$         &  \cellcolor{olive!20}\textbf{86.1}      \\
\midrule
& \multicolumn{6}{c}{{\textbf{Multi-label Classification (F1)}}$\,\uparrow$} \\
\cmidrule(l){2-7}
Mul-Net  & 70.8         & 70.9          & 74.5          & 86.2          & 83.6          & 87.7 \\
\multirow{1}{*}{TBJE}  & 68.4         & 73.9          & 74.4          & 86.3          & 83.1          & 86.6\\
\multirow{1}{*}{MR}  & 69.6         & 72.2          & 72.8          & 86.5          & 82.5          & 87.9\\
\multirow{1}{*}{COGMEN}  & \cellcolor{olive!20}\textbf{72.7}         & 73.9          & 78.0          & 86.7          & 85.5          & 88.3    \\
\multirow{1}{*}{\textsc{\textbf{Joyful}}} & 70.9         & \cellcolor{olive!20} $\textbf{\,74.6}^{\dag}$           & \cellcolor{olive!20} $\textbf{\,78.1}^{\dag}$          & \cellcolor{olive!20} $\textbf{\,89.4}^{\dag}$           & \cellcolor{olive!20} $\textbf{\,86.8}^{\dag}$           & \cellcolor{olive!20} $\textbf{\,90.5}^{\dag}$\\
\bottomrule
\end{tabular}}
\caption{Results on MOSEI with the multimodal setting. }
\label{MOSEI-main-performance}
\end{table}

Tables~\ref{performance_on_MELD} \& \ref{MOSEI-main-performance} show that \textsc{Joyful} outperformed the baselines in more complex scenes with multiple speakers or various emotional labels.
Compared with COGMEN and MM-DFN, which directly aggregate multimodal features,  \textsc{Joyful} can fully explore features from each uni-modality by specific representation learning to improve the performance.
%results for binary classification.
The GCL module can better aggregate similar emotional features for utterances to obtain better performance for multi-label classification.
We cannot improve in Happy on MOSEI since the samples are imbalanced and Happy has only 1/6 of Surprise, making \textsc{Joyful} hard to identify it. 

\begin{table}[t]
\scriptsize
\centering
\setlength{\tabcolsep}{0.7mm}{
\begin{tabular}{lcccccc}
\toprule
\multirow{2}{*}{\textit{\textbf{Modality}}} & \multicolumn{2}{c}{\textit{\textbf{IEMOCAP-4}}}                & \multicolumn{2}{c}{\textit{\textbf{IEMOCAP-6}}} & \multicolumn{2}{c}{\textit{\textbf{MOSEI (WF1)}}}
% & \multicolumn{2}{c}{\textit{\textbf{MELD}}}
\\ 
\cmidrule(lr){2-3}\cmidrule(lr){4-5}\cmidrule(l){6-7}
%\cmidrule(l){8-9}
                        & \textit{Acc.}     & \textit{WF1}       & \textit{Acc.}       & \textit{WF1}     & \textit{Binary}   & \textit{Multi-label}         
\\ 
\midrule
\midrule
Audio                     & 64.8        & 63.3          & 49.2          & 48.0          & \,\,51.2         & 53.3                 \\ 
Text%Test
& 83.0          & 83.0          & 67.4          & 67.5          & \,\,73.6          & 73.9                 \\ 
Video             & 44.6          & 43.4          & 28.2          & 28.6          & \,\,23.6          & 24.4                  \\ 
A+T                   & 82.6          & 82.5          & 67.5          & 67.8          & \,\,74.7          & 74.9                 \\ 
A+V                 & 68.0         & 67.5          & 52.7          & 52.5         & \,\,61.7          & 62.4                  \\ 
T+V               & 80.0          & 80.0          & 65.2          & 65.5      & \,\,73.1          & 73.4                  \\ 
\midrule
w/o MF(B1)                    & 85.3          & 85.4         & 70.0         & 70.3         & \,\,76.2          & 76.5                 \\
w/o MF(B2)                    & 85.2          & 85.1          & 69.2          & 69.5          & \,\,75.8          & 76.2                 \\ 
w/o MF%Fusion
& 85.2        & 84.9         & 69.0          & 69.2          & \,\,75.4         & 75.8                 \\ 
COGMEN w/o GNN  & 80.1 &80.2 &62.7 &62.9 & \,\,72.3 & 72.9 \\
w/o GCL                    & 84.7          & 84.7          & 66.1          & 66.5          & \,\,73.8          & 73.4                 \\ 
\midrule
\textsc{\textbf{Joyful}}                 & \cellcolor{olive!20}$\textbf{\,\,85.6}^{\dag}$         & \cellcolor{olive!20}$\textbf{\,\,85.7}^{\dag}$       & \cellcolor{olive!20}$\textbf{\,\,\,70.5}^{\dag}$       & \cellcolor{olive!20}$\textbf{\,\,\,71.0}^{\dag}$          & \,\,\cellcolor{olive!20}$\textbf{\,\,\,76.9}^{\dag}$          & \cellcolor{olive!20}$\textbf{\,\,77.2}^{\dag}$                 \\ 
\bottomrule
\end{tabular}}
\caption{Ablation study with different modalities.}
\label{6-way-multiple}
\end{table}

To verify the performance gain from each component, we conducted additional ablation studies.
Table~\ref{6-way-multiple} shows multi-modalities can greatly improve \textsc{Joyful}'s performance compared with each single modality.
GCL and each component of MF can separately improve the performance of \textsc{Joyful}, showing their effectiveness (Visualization in Appendix~\ref{statistical}).
%
% The two structures: 
\textsc{Joyful} w/o GCL and COGMEN w/o GNN utilize only a multimodal fusion mechanism for classification without additional modules for optimizing node representations. The comparison between them demonstrates the effectiveness of the multimodal fusion mechanism in \textsc{Joyful}.

\begin{table}[!ht]
\scriptsize
\centering
\setlength{\tabcolsep}{0.65mm}{ %Change weight of table
\begin{tabular}{cccccccc}
\toprule
\multirow{2}{*}{\textit{\textbf{Method}}} 
%& \multirow{2}{*}{\textit{\textbf{Modality}}}    
& \multicolumn{2}{c}{\textit{\textbf{One-Layer (WF1)}}}                
& \multicolumn{2}{c}{\textit{\textbf{Two-Layer (WF1)}}} 
& \multicolumn{2}{c}{\textit{\textbf{Four-Layer (WF1)}}} 
\\ 
\cmidrule(lr){2-3}\cmidrule(lr){4-5}\cmidrule(l){6-7}
&COGMEN &\textsc{Joyful} &COGMEN &\textsc{Joyful} &COGMEN &\textsc{Joyful}
\\ 
\midrule
\midrule
Unattack             
& 67.63         &71.03         & 63.21          & \cellcolor{olive!20}\textbf{71.05}          & 58.39          & 70.96                    \\
\midrule
\,~5\% Noisy        
& 65.26         & \cellcolor{olive!20}\textbf{70.82}          & 61.35          & 70.55          & 56.28          & 70.10                       \\
% \midrule
10\% Noisy           
& 62.26         & 70.33          & 59.24          & \cellcolor{olive!20}\textbf{70.45}          & 53.21          & 69.23      \\
% \midrule
15\% Noisy
% &T+A+V          
& 57.28       & \cellcolor{olive!20}\textbf{69.98}       & 55.18           & 69.21           & 52.32          
& 67.96
\\
20\% Noisy
% &T+A+V          
& 54.22       &68.52         & 51.79          & \cellcolor{olive!20}\textbf{68.82}           & 50.72         
& 67.23
\\
\bottomrule
\end{tabular}}
\caption{Adversarial attacks for GNN with different depth on 6-way IEMOCAP.}
\label{table:adversarial_attack}
\end{table} 

\begin{figure*}[t]
	\centering
	\includegraphics[scale=0.2]{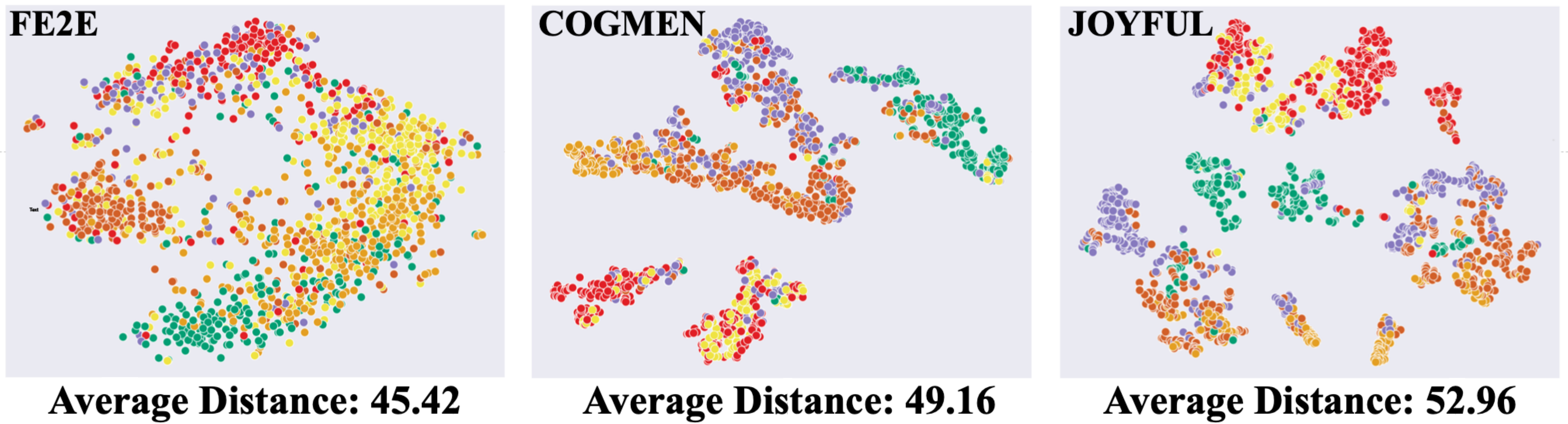}
	\caption{t-SNE visualization of IEMOCAP (6-way).}
    \label{t-SNE_visualization}
\end{figure*}
We deepened the GNN layers to verify \textsc{Joyful}'s ability to alleviate the over-smoothing. 
In Table~\ref{table:adversarial_attack}, COGMEN with four-layer GNN was 9.24\% lower than that with one-layer, demonstrating that the over-smoothing decreases performance, 
while \textsc{Joyful} relieved this issue by using the GCL framework.  
To verify the robustness, following \citet{TF-GCL},  we randomly added 5\%$\sim$20\% noisy edges to the training data.
In Table~\ref{table:adversarial_attack}, COGMEN was easily affected by the noise, decreasing 10.8\% performance in average with 20\% noisy edges, 
while \textsc{Joyful} had strong robustness with only an average 2.8\% performance reduction for 20\% noisy edges.

\begin{figure}[t]
	\centering
	\includegraphics[scale=0.42]{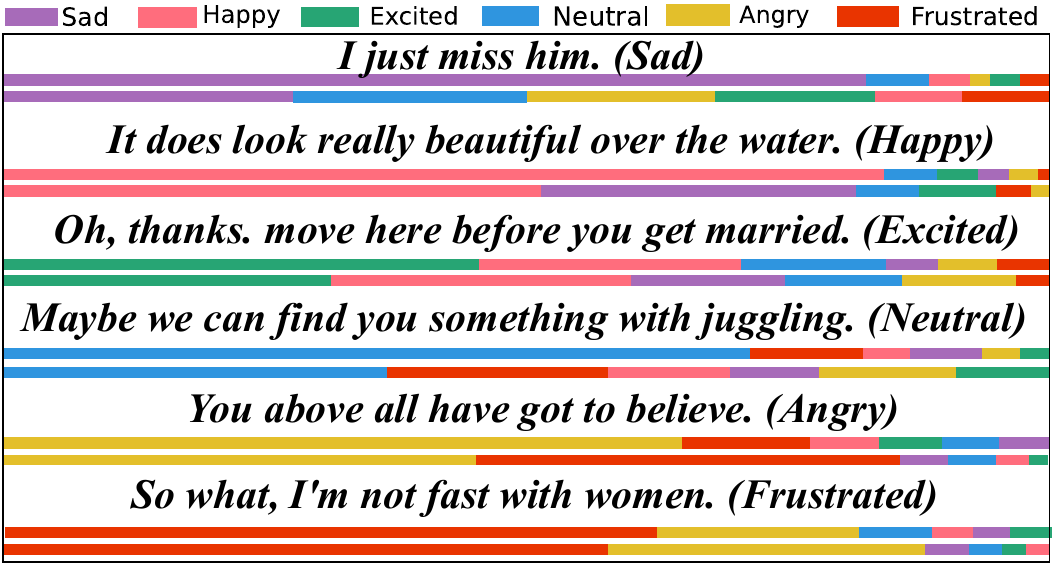}
	\caption{Visualization of emotion probability, each first row is \textsc{Joyful} and each second row is COGMEN.}
    \label{fig:emb_visualization}
\end{figure}

% \subsection{Case Study}
To show the distinguishability of the node representations, we visualize the node representations of FE2E, COGMEN, and \textsc{Joyful} on 6-way IEMOCAP. 
In Figure~\ref{t-SNE_visualization}, COGMEN and \textsc{Joyful} obtained more distinguishable node representations than FE2E, demonstrating that graph structure is more suitable for MERC than Transformers. 
\textsc{Joyful} performed better than COGMEN, illustrating the effectiveness of GCL.
In Figure~\ref{fig:emb_visualization}, we randomly sampled one example from each emotion of IEMOCAP (6-way) and chose best-performing COGMEN for comparison.
\textsc{Joyful} obtained more discriminate prediction scores among emotion classes, showing GCL can push samples from different emotion class farther apart.

\section{Conclusion}

We proposed a joint learning model (\textsc{Joyful}) for MERC, that involves a new multimodal fusion mechanism and GCL module to effectively improve the performance of MERC.
The MR mechanism can extract and fuse contextual and uni-modal specific emotion features, and the GCL module can help learn more distinguishable representations.
For future work, we plan to investigate the performance of using supervised GCL for \textsc{Joyful} on unbalanced and small-scale emotional datasets.

\section*{Acknowledgements}

The authors would like to thank Ying Zhang~\footnote{\href{https://scholar.google.com/citations?user=tbDNsHsAAAAJ&hl=en}{scholar.google.com/citations?user=tbDNsHs}} for her advice and assistance. 
We gratefully acknowledge anonymous reviewers for their helpful comments and feedback. 
We also acknowledge the authors of COGMEN~\cite{joshi-etal}: Abhinav Joshi and Ashutosh Modi for sharing codes and datasets.
Finally, Dongyuan Li acknowledges the support of the China Scholarship Council (CSC).

\section*{Limitations}
\textsc{Joyful} has a limited ability to classify minority classes with fewer samples in unbalanced datasets.
Although we utilized self-supervised graph contrastive learning to learn a distinguishable representation for each utterance by exploring vertex attributes, graph structure, and contextual information, GCL failed to separate classes with fewer samples from the ones with more samples because the utilized self-supervised learning lacks the label information and does not balance the label distribution.
Another limitation of \textsc{Joyful} is that its framework was designed specifically for multimodal emotion recognition tasks, which is not straightforward and general as language models~\citep{DevlinCLT19,abs-1907-11692} or image processing techniques~\citep{lecun1995convolutional}. 
This setting may limit the applications of \textsc{Joyful} for other multimodal tasks, such as the multimodal sentiment analysis task (Detailed experiments in Appendix~\ref{MSA}) and the multimodal retrieval task. 
Finally, although \textsc{Joyful} achieved SOTA performances on three widely-used MERC benchmark datasets, its performance on larger-scale and more heterogeneous data in real-world scenarios is still unclear.

\bibliography{anthology}
\bibliographystyle{acl_natbib}

\clearpage
\appendix

% \section{Example of Graph Construction}
% \label{appendix:graph_construction}

% Graph construction aims to establish relations between past and future utterances.
% In Figure~\ref{node_example}, \textit{Intra-Dependency} ($\mathcal{R}_{intra}\in \{\mathcal{U}^{S_{i}} \to \mathcal{U}^{S_{i}}\}$) represents intra-relations between the utterances (red lines), 
% and \textit{Inter-Dependency} ($\mathcal{R}_{inter} \in \{\mathcal{U}^{S_{i}} \to \mathcal{U}^{S_{j}}\}, i \neq j$) represents the inter-relations between the utterances (purple lines). 
% And we show that window size is a hyper-parameter that controls the context information for each utterance and provide accurate intra- and inter-speaker dependencies.

% \begin{figure}[!ht]
% \centering
% \includegraphics[width=0.47\textwidth]{Figure/parameter_tuning_new_6.pdf}
% \caption{An example of graph construction.}
% \label{node_example}
% \end{figure}

\section{Example for Global Proximity}
\label{High-order}

\begin{figure}[!ht]
\includegraphics[width=0.42\textwidth]{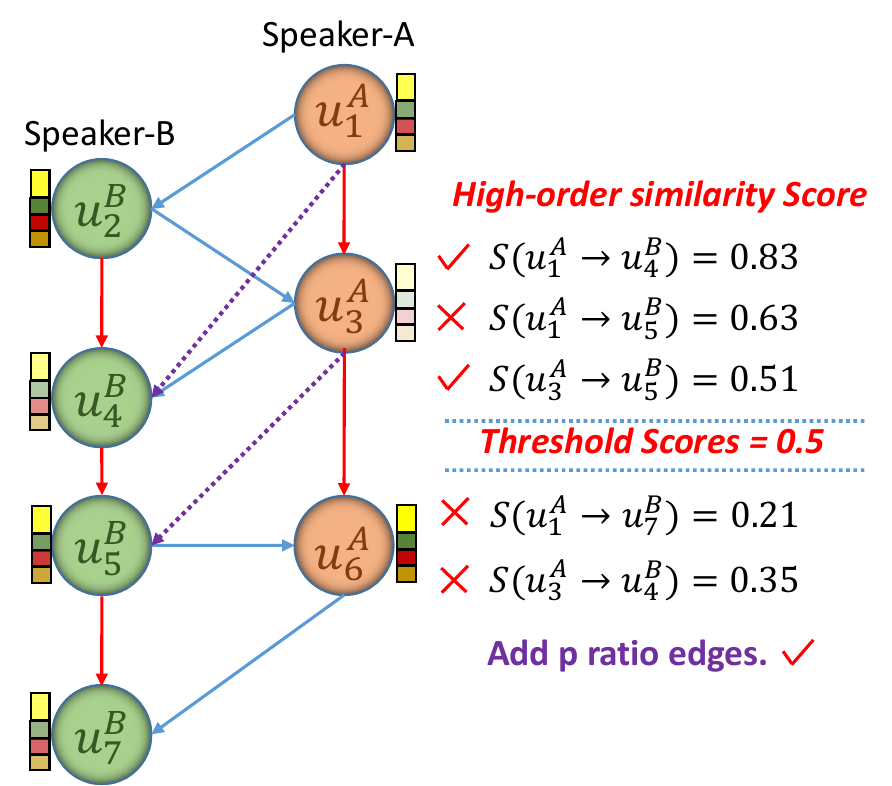}
\centering
\caption{Example of adding $p$\% high-order edges to explore global topological information of graph.}
\label{high_example}
\end{figure}

In Figure~\ref{high_example}, given the network $\mathcal{G}$ and a modified $p$, we first used the Katz index~\citep{katz1953new} to calculate a high-order similarity between the vertices.
We considered the arbitrary number of high-order distances. 
For example, second-order similarity between $u_{1}^{A}$ and $u_{4}^{B}$ as $u_{1}^{A} \to u_{4}^{B} =0.83 $,  third-order similarity between $u_{1}^{A}$ and $u_{5}^{B}$ as $u_{1}^{A} \to u_{5}^{B} =0.63 $, and fourth-order similarity between $u_{1}^{A}$ and $u_{7}^{B}$ as $u_{1}^{A} \to u_{7}^{B} =0.21 $.
We then define the threshold score as 0.5, where a high-order similarity score less than the threshold will not be selected as added edges. 
Finally, we randomly selected $p$\% edges (whose scores are higher than the threshold score) and added them to the original graph $\mathcal{G}$ to construct the augmented graph. 

\section{Dimensions of Mathematical Symbols}
\label{Dimension}
Since we do not have much space to introduce details about the dimensions of the mathematical symbols in our main body. 
We carefully list all the dimensions of the mathematical symbols of IEMOCAP in Table~\ref{Dimensions-1}. 
Mathematical symbols for other two datasets please see our source code. 
% and Table~\ref{Dimensions-2} to make our equations more easily to understand (More details can see our source code).

%\begin{scriptsize}
\begin{table}[!ht]\scriptsize
\centering
\setlength{\tabcolsep}{0.2mm}{ %Change weight of table
% \footnotesize
\begin{tabular}{lc}
\toprule
\textit{\textbf{Symbols}}     
& \multicolumn{1}{c}{\textbf{\textit{Description}}}  \\ 
\midrule 
\midrule
$\bm{x}_{v} \in \mathbb{R}^{512}$   & Video Features   \\ 
$\bm{x}_{a} \in \mathbb{R}^{100}$   & Audio Features   \\ 
$\bm{x}_{t} \in \mathbb{R}^{768}$   & Text Features   \\
\midrule
\multicolumn{2}{c}{\textcolor{orange}{Contextual Representation Learning}}\\
\midrule
$\bm{z}_{v}^{g} \in \mathbb{R}^{512}$   & Global Hidden Video Features   \\
$\bm{z}_{a}^{g} \in \mathbb{R}^{100}$   & Global Hidden Audio Features   \\
$\bm{z}_{t}^{g} \in \mathbb{R}^{768}$   & Global Hidden Text Features   \\
$\bm{z}_{m}^{g} \in \mathbb{R}^{1{,}380}$   & Global Combined Features   \\
$\bm{z}_{con} \in \mathbb{R}^{1,380}$   & Topic-related Vector   \\
$\bm{\hat{z}}_{m}^{g} \in \mathbb{R}^{1,380}$   & Global Output Features   \\
\midrule
\multicolumn{2}{c}{\textcolor{orange}{Specific Representation Learning}}\\
\midrule
$\bm{z}_{v}^{\ell} \in \mathbb{R}^{460}$   & Local Hidden Video Features   \\
$\bm{z}_{a}^{\ell} \in \mathbb{R}^{460}$   & Local Hidden Audio Features   \\
$\bm{z}_{t}^{\ell} \in \mathbb{R}^{460}$   & Local Hidden Text Features   \\
$\bm{b}_{m} \in \mathbb{R}^{460}$   & Basic Features   \\
$ \tilde{\bm{z}}_{\{v,a,t\}}^{\ell} \in \mathbb{R}^{460}$   & Features in Shared Space   \\
$ \tilde{\bm{b}}_{m} \in \mathbb{R}^{460}$   & Basic Features in Shared Space   \\
$ \textbf{W}_{\{v,a,t,b\}} \in \mathbb{R}^{460\times 460}$   & Trainable Matrices   \\
$\hat{\bm{z}}_{\{v,a,t\}}^{\ell} \in \mathbb{R}^{460}$   & New Multimodal Features   \\
$\hat{\bm{z}}_{m}^{\ell} \in \mathbb{R}^{1,380}$   & New Multimodal Combined Features   \\
$\bm{z}_{m}^{\ell} \in \mathbb{R}^{1380}$   & Original Combined Features   \\
\midrule
\multicolumn{2}{c}{\textcolor{orange}{Graph Contrastive Learning (One GCN Layer) }}\\
\midrule
$(\hat{\bm{z}}_{g}^{\ell} \| \hat{\bm{z}}_{m}^{\ell}) \in \mathbb{R}^{2,760}$   &  Global-Local Combined Features  \\
$\textrm{AGG} \in \mathbb{R}^{2,760 \times 2,760}$ & Parameters of Aggregation Layer\\
$\textrm{COM} \in \mathbb{R}^{2,760 \times 5,520}$ & Input/Output of Combination Layer\\
$\textbf{W}_{graph} \in \mathbb{R}^{5,520 \times 2,760}$ & Dimention Reduction after COM \\
$\bm{h}_{m} \in \mathbb{R}^{2,760}$ & Node Features of GCN Layer\\
\bottomrule
\end{tabular}}
\caption{Mathematical symbols for IEMOCAP dataset.}
\label{Dimensions-1}
\end{table}
%\end{scriptsize}

\begin{figure*}[!ht]
\includegraphics[width=1\textwidth]{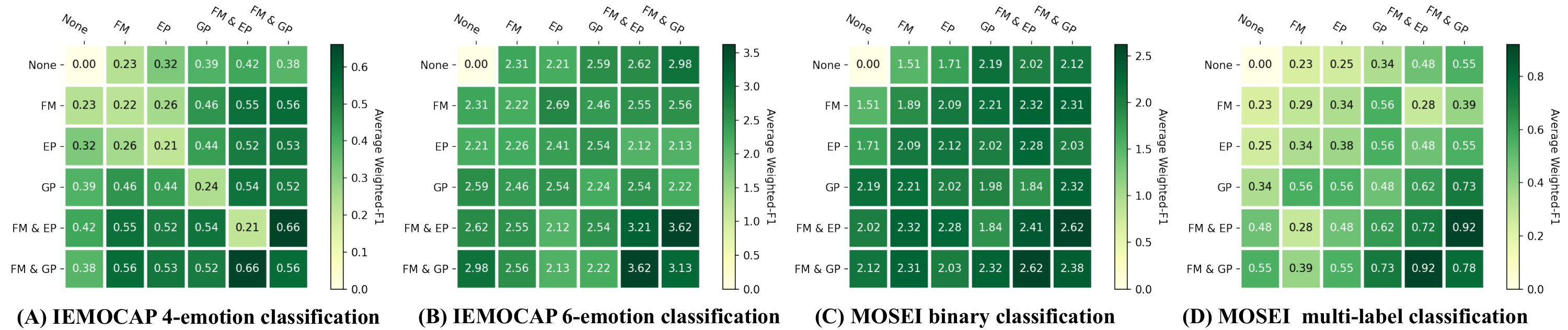}
\centering
\caption{Average WF1 gain when contrasting different augmentation pairs, compared with training without graph augmentation module.}
\label{fig:aug_compare} 
\end{figure*}

\section{Observations of Graph Augmentation}
\label{Data_augmentation}

As shown in Figure~\ref{fig:aug_compare},
when we consider the combinations of (FM $\&$ EP) and (FP $\&$ GP) as two graph augmentation methods of the original graph, we could achieve the best performance. 
Furthermore, we have the following observations:

\paragraph{\small{Obs.1: Graph augmentations are crucial.}}
Without any data augmentation, GCL module will not improve accuracy, judging from the averaged WF1 gain of the pair (None, None) in the upper left corners of Figure~\ref{fig:aug_compare}. 
In contrast, composing an original graph and its appropriate augmentation can benefit the averaged WF1 of emotion recognition, judging from the pairs (None, any) in the top rows or the left-most columns of Figure~\ref{fig:aug_compare}.
Similar observation were in graphCL~\cite{YouCSCWS20},
without augmentation, GCL simply compares two original samples as a negative pair with the positive pair loss becoming zero, which leads to homogeneously pushes all graph representations away from each other. 
Appropriate augmentations can enforce the model to learn representations invariant to the desired perturbations through maximizing the agreement between a graph and its augmentation.

\paragraph{\small{Obs.2: Composing different augmentations benefits the model's performance more.}} 
Applying augmentation pairs of the same type does not often result in the best performance (see diagonals in Figure~\ref{fig:aug_compare}). 
In contrast, applying augmentation pairs of different types result in better performance gain (see off-diagonals of Figure~\ref{fig:aug_compare}).
Similar observations were in SimCSE~\citep{nlp2}. 
As mentioned in that study, composing augmentation pairs of different types correspond to a ``harder'' contrastive prediction task, which could enable learning more generalizable representations.

\paragraph{\small{Obs.3: One view having two augmentations result in better performance.}} 
Generating each view by two augmentations further improve performance, i.e., the augmentations FM $\&$ EP, or FM $\&$ GP.
The augmentation pair (FM $\&$ EP, FM $\&$ GP) results in the largest performance gain compared with other augmentation pairs. 
We conjectured the reason is that simultaneously changing structural and attribute information of the original graph can obtain more heterogeneous contextual information for nodes, which can be consider as ``harder'' example to prompt the GCL model to obtain more generalizable and robust representations.

\begin{table}[!ht]
\centering
\scriptsize
\setlength{\tabcolsep}{1.3mm}{ %Change weight of table
\begin{tabular}{lcccccc}
\toprule
\multirow{1}{*}{\textit{\textbf{P\&F}}} & \textit{Happiness}     & \textit{Sadness}       & \textit{Neutral}       & \textit{Anger}     & \textit{Accuracy}   & \textit{WF1}          \\ 
\midrule
\midrule
size=1   &83.27 &83.04 &80.63 &81.54 &81.87 &81.82 \\
size=2   &79.02 &82.92 &83.93 &86.65 &83.46 &83.41               \\ 
size=3   &80.88 &86.34 &84.07 &85.64 &84.52 &84.45            \\ 
size=4   &\textbf{83.92} &{85.83} &83.91 &84.35 &84.52 &84.51              \\ 
size=5   &82.93 &87.85 &83.79 &86.47 &85.26 &85.20              \\ 
\rowcolor{yellow-green!30}
size=6   &81.73 &86.42 &85.17 &88.46 &85.58 &85.56                \\ 
\rowcolor{yellow-green!30}
size=7   &79.33 &86.07 &83.29 &86.40 &83.99 &83.97               \\ 
\rowcolor{Ocean!30} size=8   &80.14 &\textbf{88.11} &85.06 &88.15 &\textbf{85.68} &\textbf{85.66}                 \\ 
size=9   &77.29 &87.85 &83.56 &87.19 &84.41 &84.37              \\ 
size=10  &80.00 &87.47 &\textbf{85.29} &\textbf{88.64} &85.68 &85.66                  \\ 
size=ALL &79.87 &84.35 &83.20 &84.75 &83.24 &83.24                  \\ 
\bottomrule
\end{tabular}}
\caption{Results for various window sizes for graph formation on the IEMOCAP (4-way).}
\label{4-way-window_size}
\end{table}

\begin{table}[!ht]
\scriptsize
\centering
\setlength{\tabcolsep}{1.2mm}{ %Change weight of table
\begin{tabular}{lcccccccc}
\toprule

\multirow{1}{*}{\textit{\textbf{P\&F}}}
                        & \textit{Hap.}     & \textit{Sad.}       & \textit{Neu.}       & \textit{Ang.}
                        & \textit{Exc.} & \textit{Fru.} & \textit{Acc.}   & \textit{WF1}          \\ 
\midrule
\midrule
size=1  &57.85 &80.43 &62.88 &60.61 &70.76 &60.99 &65.50 &65.85\\ 
size=2  &56.27 &79.57 &64.17 &60.87 &72.50 &61.52 &65.93 &66.36           \\ 
size=3  &60.80 &80.26 &66.06 &64.47 &73.17 &62.70 &67.71 &68.09       \\ 
size=4  &59.95 &80.79 &67.96 &67.18 &71.60 &64.89 &68.64 &69.05        \\ 
size=5  &60.06 &81.42 &68.23 &66.33 &73.88 &63.24 &68.76 &69.17      \\ 
\rowcolor{Ocean!30}
size=6  &\textbf{60.94} &\textbf{84.42} &68.24 &\textbf{69.95} &73.54 &\textbf{67.55} &\textbf{70.55} &\textbf{71.03}   \\ 
\rowcolor{yellow-green!30}
size=7  &59.84 &80.53 &67.93 &68.12 &73.72 &63.91 &68.82 &69.26      \\ 
\rowcolor{yellow-green!30}
size=8  &57.66 &82.17 &\textbf{70.56} &67.53 &73.92 &64.79 &69.75 &70.12           \\ 
size=9  &58.01 &81.13 &70.22 &65.42 &\textbf{75.05} &61.49 &68.82 &69.12      \\ 
size=10 &59.77 &81.84 &69.17 &65.85 &73.56 &63.51 &68.95 &69.38            \\ 
size=ALL &54.74 &78.75 &66.58 &64.56 &68.63 &63.46 &66.42 &66.80        \\ 
\bottomrule
\end{tabular}}
\caption{Results for various window sizes for graph formation on the IEMOCAP (6-way).}
\label{6-way-window_size}
\end{table}

\section{Parameters Sensitivity Study}
\label{parameters}

In this section, we give more details about parameter sensitivity. 
First, as shown in Tables~\ref{4-way-window_size} \& \ref{6-way-window_size},  when the window size $\in [6,8]$ for IEMOCAP (6-way) and the window size is 6 for IEMOCAP (4-way), \textsc{Joyful} achieved the best performance. 
A small window size will miss much contextual information, and a large-scale window size contains too much noise (topic will change over time).
We set the window size for past and future to 6.

\begin{figure*}[!ht]
\includegraphics[width=1\textwidth]{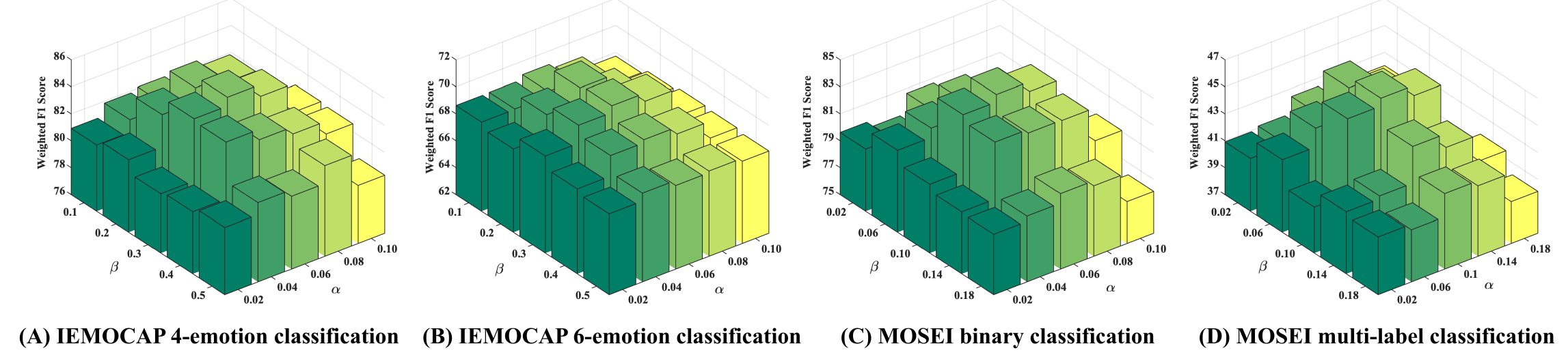}
\centering
\caption{Parameters tuning for $\boldsymbol{\alpha}$ and $\boldsymbol{\beta}$ on validation datasets for all multimodal emotion recognition tasks.}
\label{fig:parameter_all}
\end{figure*}

% \begin{figure*}[htb!]
% 	\centering
% 	\includegraphics[scale=0.7]{Figure/Ablation_all.pdf}
%     \label{fig_random}
% \end{figure*}

\textsc{Joyful} also has two hyper-parameters: $\boldsymbol{\alpha}$ and $\boldsymbol{\beta}$,  which balance the importance of MF module and GCL module in Eq.(\ref{cmall}).
Specifically, as shown in Figure~\ref{fig:parameter_all}, we observed how $\boldsymbol{\alpha}$ and $\boldsymbol{\beta}$ affect the performance of \textsc{Joyful} by varying $\boldsymbol{\alpha}$ from 0.02 to 0.10 in 0.02 intervals and $\boldsymbol{\beta}$ from 0.1 to 0.5 in 0.1 intervals.
The results indicate that \textsc{Joyful} achieved the best performance when $\boldsymbol{\alpha} \in [0.06,0.08]$ and $\boldsymbol{\beta} \in [0.2,0.3]$ on IEMOCAP and 
and when $\boldsymbol{\alpha} \in [0.06,0.1]$ and $\boldsymbol{\beta} =0.1 $ on MOSEI.
The reason why these parameters can affect the results is that when $\boldsymbol{\alpha}$< 0.06, 
MF becomes weaker and representations contain too much noise, which cannot provide a good initialization for downstream MERC tasks. 
When $\boldsymbol{\alpha}$ >0.1,  it tends to make reconstruction loss more important and \textsc{Joyful} tends to extract more common features among multiple modalities and loses attention to explore features from uni-modality. 
When $\boldsymbol{\beta}$ is small, graph contrastive loss becomes weaker, which leads to indistinguishable representation. 
A larger $\boldsymbol{\beta}$  wakes the effect of MF,  leading to a local optimal solution. 
We set $\boldsymbol{\alpha}$=0.06 and $\boldsymbol{\beta}$=0.3 for IEMOCAP and MELD. 
We set $\boldsymbol{\alpha}$=0.06  and $\boldsymbol{\beta}$ =0.1 for MOSEI.

\begin{table}[!ht]
\footnotesize
\centering
\setlength{\tabcolsep}{1mm}{ %Change weight of table
\begin{tabular}{lcc}
\toprule

\multirow{1}{*}{\textit{\textbf{Method}}}
                        & \textit{Modality}     & \textit{WF1}            \\ 
\midrule
\midrule
\multicolumn{3}{c}{\textit{\textbf{IEMOCAP 6-way}}} \\
\midrule
\midrule
% EmoBERTa   &Text &\cellcolor{orange!30}\textbf{68.57}  \\
CESTa   &Text &67.10           \\ 
SumAggGIN   &Text &66.61           \\ 
DiaCRN   &Text &66.20           \\ 
DialogXL   &Text &65.94           \\ 
DiaGCN   &Text &64.18           \\ 
COGMEN   &Text &66.00           \\
\cellcolor{yellow-green!30}{DAG-ERC}   &Fine-tune Text (RoBERTa-large) &\cellcolor{yellow-green!30}68.03              \\ 
\midrule
\multirow{3}{*}{\textit{\textbf{\textsc{Joyful}}}}  
&Text (Sentence-BERT) &67.48           \\
&Text (RoBERTa-large) &68.05\\
&Fine-tune Text (RoBERTa-large) & \cellcolor{Ocean!30}\textbf{68.45} \\
&A+T+V &\cellcolor{Ocean!30}\textbf{71.03}           \\
\bottomrule
\end{tabular}}
\caption{Overall performance comparison on MOSEI with Text Modality.}
\label{4-unimodal}
\end{table}

\section{Uni-modal Performance}
\label{Uni-modal_performance}
The focus of this study was multimodal emotion recognition. 
However, we also compared \textsc{Joyful}
with uni-modal methods to evaluate its performance of \textsc{joyful}. 
We compared it with 
DAG-ERC~\citep{ShenWYQ20}, CESTa~\citep{wang-etal-2020-contextualized}, 
SumAggGIN~\citep{sheng-etal-2020-summarize}, DiaCRN~\citep{hu-etal-2021-dialoguecrn}, DialogXL~\citep{ShenCQX21}, DiaGCN~\citep{dialoguegcn}, and COGMEN~\citep{joshi-etal}.
Following COGMEN, 
text-based models were specifically optimized for text modalities and incorporated changes to architectures to cater to text.
As shown in Table~\ref{4-unimodal}, \textsc{Joyful}, being a fairly generic architecture, still achieved better or comparable performance with respect to the state-of-the-art uni-modal methods. 
Adding more information via other modalities helped to further improve the performance of \textsc{Joyful} (Text vs A+T+V). 
When using only text modality, the DAG-ERC baseline could achieve higher WF1 than \textsc{Joyful}. 
And we conjecture the main reasons is:
% \begin{quote}
DAG-ERC~\citep{ShenWYQ20} fine-tuned RoBERTa large model~\citep{abs-1907-11692}, with 354 million parameters, as their text encoder. 
By fine-tuning on RoBERTa large model under the guidance of downstream emotion recognition signals,  RoBERTa large model can provide the most suitable text features for ERC. 
Compared with DAG-ERC, \textsc{Joyful} and other methods directly use Sentence-BERT~\citep{2019-sentence}, with 110 million parameters, as their text encoder without fine-tuning on ERC datasets.
% \end{quote}

To verify whether the above inference is reasonable, we used RoBERTa large model as our text feature extractor called \textit{Text (RoBERTa-large)}. 
And we fine-tuned RoBERTa large model on the downstream IEMPCAP (6-way) dataset, following the same method of DAG-ERC called \textit{Fine-tune Text (RoBERTA-large)}.
The observation meets our intuition. 
With RoBERTa large model, \textsc{Joyful} improved the performance (68.05 vs 67.48) compared with Sentence-BERT as our text encoder.
And \textsc{Joyful} could obtain better performance (68.45 vs 68.03) in terms of WF1 than DAG-ERC with fine-tuned RoBERTa-large, demonstrating that fine-tuning large-scale model can help obtain richer text features to improve the performance.
However, considering a fair comparison with other multimodal emotion recognition baselines (they do not have the fine-tuning process~\citep{joshi-etal,dialoguegcn}) and saving the additional time-consuming on fine-tuning, we directly adopt Sentence-BERT as our text encoder for IEMOCAP.

\section{Pseudo-Code of \textsc{Joyful}}
\label{pseudo}

As shown in Algorithm~\ref{alg:overall_process}, to make \textsc{Joyful} easy to understand, we also provide a pseudo-code. 

% \begin{footnotesize}
% \IncMargin{0em}
% \setlength{\tabcolsep}{0.01mm}{ %Change weight of table
\begin{algorithm}\footnotesize
% \scriptsize
\SetKwData{Left}{left}\SetKwData{This}{this}\SetKwData{Up}{up} \SetKwFunction{Union}{Union}\SetKwFunction{FindCompress}{FindCompress} \SetKwInOut{Input}{input}\SetKwInOut{Output}{output}
\Input{Visual features $\bm{x}_{v}$;\\
Audio features $\bm{x}_{a}$;\\
Text features $\bm{x}_{t}$;\\
Parameters: $\boldsymbol{\alpha}$, $\boldsymbol{\beta}$, Window size} 
\Output{Emotion recognition label.}
\BlankLine 
Initialize trainable parameters\; 
\For{$epoch\leftarrow 1$ \KwTo $epoch \ num$}{
Global Contextual Fusion $\hat{\bm{z}}_{m}^{g}$\; 
%= ($\bm{z}_{v}^{g}\|\bm{z}_{a}^{g}\|\bm{z}_{t}^{g}$)\;
Specific Modality Fusion $\hat{\bm{z}}_{m}^{\ell}$=($\bm{z}_{v}^{g}\|\bm{z}_{a}^{g}\|\bm{z}_{t}^{g}$)\;
\tcp{\scriptsize{\textcolor{blue}{\textbf{Compute multimodal fusion loss}}}}
Compute $\mathcal{L}_{mf}$, in accordance with Eq.(\ref{cm1})\;
Feature Concatenation $\bm{h}= (\hat{\bm{z}}_{m}^{g} \| \hat{\bm{z}}_{m}^{\ell}$)\;
Adopt $\bm{h}$ as initialization for Graph\;
\tcp{\scriptsize{\textcolor{blue}{\textbf{Generate two augmented views}}}}
Apply FM \& EP to generate view: $\mathcal{G}^{(1)}$\; 
Apply FM \& GP to generate view: $\mathcal{G}^{(2)}$\;
\tcp{\scriptsize{\textcolor{blue}{\textbf{Extract features of two views}}}}
$\mathbf{H}^{(1)}$ = $GCNs(\mathcal{G}^{(1)})$, 
$\mathbf{H}^{(2)}$ = $GCNs(\mathcal{G}^{(2)})$ \;
\tcp{\scriptsize{\textcolor{blue}{\textbf{Compute contrastive learning loss}}}}
Compute $\mathcal{L}_{ct}$, in accordance with Eq.(\ref{cm2}) \;
\tcp{\scriptsize{\textcolor{blue}{\textbf{Aggregate extracted features}}}}
$\mathbf{H} = \mathbf{H}^{(1)} + \mathbf{H}^{(2)}$ \;
\tcp{\scriptsize{\textcolor{blue}{\textbf{Compute emotion recognition loss}}}}
Compute $\mathcal{L}_{ce}$, in accordance with Eq.(\ref{cm3})\;
\tcp{\scriptsize{\textcolor{blue}{\textbf{Joint training}}}}
Compute $\mathcal{L}_{all}$, in accordance with
Eq.(\ref{cmall})\;
\tcp{\scriptsize{\textcolor{blue}{\textbf{Optimize with Adam optimizer}}}}
}
Adopt classifier on $\mathbf{H}$ to predict the emotional label.
\caption{Overall process of \textsc{Joyful}}
\label{alg:overall_process}
\end{algorithm}
% \DecMargin{0em}
% \end{footnotesize}

\section{Benjamini-Hochberg Correction} 
\label{B-H_analysis}

Benjamini-Hochberg Correction (\textit{B-H})~\citep{benjam} is a powerful tool that decreases the false discovery rate.
Considering the reproducibility of the multiple significant test, we introduce how we adopt the \textit{B-H} correction and give the hyper-parameter values that we used.
We first conduct a t-test~\citep{test} with default parameters\footnote{\href{https://docs.scipy.org/doc/scipy/reference/generated/scipy.stats.ttest_ind.html}{scipy.stats.ttest\_ind.html}} to calculate the p-value between each comparison method with \textsc{Joyful}.
We then put the individual p-values in ascending order as input to calculate the p-value corrected using the \textit{B-H} correction. 
We directly use the ``\textit{\textbf{multipletests(*args)}}'' function from python package\footnote{\href{https://www.statsmodels.org/dev/generated/statsmodels.stats.multitest.multipletests.html}{statsmodels.stats.multitest.multipletests.html}} and set the hyperparameter of the false discovery rate $Q = 0.05$, which is a widely used default value~\citep{PUOLIVALI}. 
Finally, we obtain a cut-off value as the output of the \textbf{\textit{multipletests}} function, where cut-off is a dividing line that distinguishes whether two groups of data are significant.
If the p-value is smaller than the cut-off value, we can conclude that two groups of data are significantly different.
%
% \begin{figure}[ht]
% \includegraphics[width=0.45\textwidth]{Figure/significant.pdf}
% \centering
% \caption{Examples of statistical analysis of comparison methods. 
% The p-value between \textsc{Joyful} and compared methods are less than 0.005.}
% \label{fig:significant}
% \end{figure}
\begin{figure*}[t]
\includegraphics[width=0.8\textwidth]{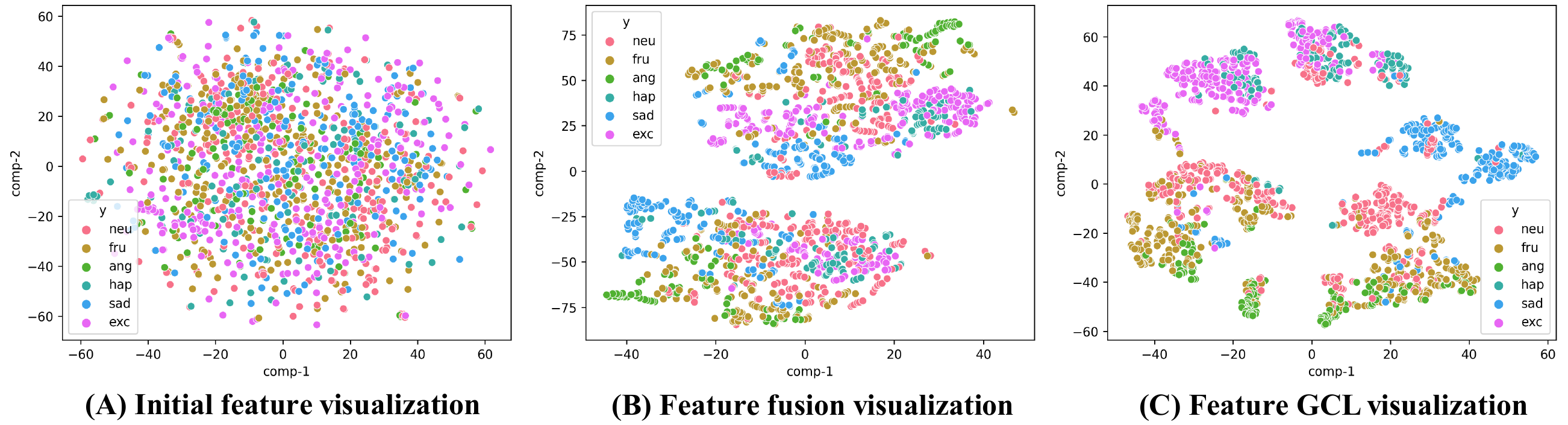}
\centering
\caption{t-SNE visualization of IEMOCAP (6-way) features.}
\label{fig:visualization_full}
\end{figure*}
% As shown in Figure~\ref{fig:significant}, we give IEMOCAP (4-way \& 6-way) significant analysis. 
% For IEMOCAP (4-way), we chose MFN, COGMEN, MTAG and MICA as our comparison methods since they can achieve the best performance. 
% The p-values between \textsc{Joyful} and MFN, COGMEN, MTAG and MICA were $1.64e^{-90}$, 
% $6.55e^{-56}$, $2.79e^{-82}$, and $3.88e^{-108}$, respectively. 
% For IEMOCAP (6-way), we chose MFN, COGMEN, Af-CAN, and RGAT as our comparison methods. 
% The p-values between \textsc{Joyful} and MFN, COGMEN, Af-CAN, and RGAT were $4.9e^{-108}$, 
% $3.04e^{-56}$, $6.07e^{-77}$, and $3.88e^{-70}$, respectively. 
% All p-values are much less than 0.005, which indicates that \textsc{Joyful} significantly outperformed all other methods.

The use of t-test for testing statistical significance may not be appropriate for F-scores, as mentioned in \citet{dror-etal-2018-hitchhikers}, as we cannot assume normality.
To verify whether our data meet the normality assumption and the homogeneity of variances required for the t-test, following \citet{shapiro1965analysis} and \citet{levene1960contributions}, we conducted the following validation. 
First, we performed the Shapiro-Wilk test on each group of experimental results to determine whether they are normally distributed. Under the constraint of a significance level (alpha=0.05), all p-values resulting from the Shapiro-Wilk test~\footnote{\href{https://docs.scipy.org/doc/scipy/reference/generated/scipy.stats.shapiro.html}{scipy.stats.shapiro.html}} for the baselines and our model were greater than 0.05. 
This indicates that the results of the baselines and our model all adhere to the assumption of normality. For example, in IEMOCAP-4, p-values for [Mult, RAVEN, MTAG, PMR, MICA, COGMEN, JOYFUL] are [0.903, 0.957, 0.858, 0.978, 0.970, 0.969, 0.862].
Furthermore, we used the Levene's test~\cite{schultz1985levene} to check for homogeneity of variances between baselines and our model. Under the constraint of a significance level (alpha = 0.05), we found that our p-values are greater than 0.05, indicating the homogeneity of the variances between the baselines and our model. For example, we obtained p-values 0.3101 and 0.3848 for group-based baselines on IEMOCAP-4 and IEMOCAP-6, respectively.
Since we were able to demonstrate that all baselines and our model conform to the assumptions of normality and homogeneity of variances, we believe that the significance tests we reported are accurate.

\section{Representation Visualization}
\label{statistical}

We visualized the node features to understand the function of the multimodal fusion mechanism and the GCL-based node representation learning component, as shown in Figure~\ref{fig:visualization_full}. 
Figure~\ref{fig:visualization_full} (A) shows the concatenated multimodal features on the input side. 
Figure~\ref{fig:visualization_full} (B) shows the representation of utterances after the feature fusion module.
Figure~\ref{fig:visualization_full} (C) shows the representation of the utterances after the GCL module (Eq.(10)) and before the pre-softmax layer (Eq.(11)). 
We observed that utterances could be roughly separated after the feature fusion mechanism, which indicates that the multimodal fusion mechanism can learn distinctive features to a certain extent.  
After GCL-based module, \textsc{Joyful} can be easily separated, demonstrating that GCL can provide distinguishable representation by exploring vertex attributes, graph structure, and contextual information from datasets.

\section{Labels Distribution of Datasets}
\label{label_distribution}

In this section, we list the detailed label distribution of the three multimodal emotion recognition datasets MELD (Table~\ref{data_1}), IEMOCAP 4-way (Table~\ref{data_2}), IEMOCAP 6-way (Table~\ref{data_3}) and MOSEI (Table~\ref{data_4}) in the draft. 

\begin{table}[ht]
\centering
\footnotesize
\setlength{\tabcolsep}{4mm}{ %Change weight of table
\begin{tabular}{lccc}
\toprule
\textit{\textbf{MELD}}     
& \multicolumn{1}{c}{Train} & \multicolumn{1}{c}{\,\,\,\,Valid} & Test \\ 
\midrule 
\midrule
\multicolumn{1}{l}{Anger}    & \multicolumn{1}{c}{1,109}  
& \multicolumn{1}{c}{153}   & \multicolumn{1}{c}{345}  \\ 
\multicolumn{1}{l}{Disgusted}    & \multicolumn{1}{c}{271}  
& \multicolumn{1}{c}{22}  & \multicolumn{1}{c}{68} \\ 
% \midrule
\multicolumn{1}{l}{Fear}  & \multicolumn{1}{c}{268}
& \multicolumn{1}{c}{40}  & \multicolumn{1}{c}{50}\\ 
\multicolumn{1}{l}{Joy}  & \multicolumn{1}{c}{1,743} 
& \multicolumn{1}{c}{163}  & \multicolumn{1}{c}{402} \\ 
\multicolumn{1}{l}{Neutral}  & \multicolumn{1}{c}{4,710} 
& \multicolumn{1}{c}{470}  & \multicolumn{1}{c}{1,256} \\ 
\multicolumn{1}{l}{Sadness}  & \multicolumn{1}{c}{683}
& \multicolumn{1}{c}{111}  & \multicolumn{1}{c}{208} \\ 
\multicolumn{1}{l}{Surprise}  & \multicolumn{1}{c}{1,205}
& \multicolumn{1}{c}{150}  & \multicolumn{1}{c}{281} \\
\midrule
\multicolumn{1}{l}{Total}  & \multicolumn{1}{c}{9,989}
& \multicolumn{1}{c}{1,109}  & \multicolumn{1}{c}{2,610} \\ 
\bottomrule
\end{tabular}}
\caption{Labels distribution of MELD dataset.} 
%IEMOCAP shows the number of utterances/conversations. }
\label{data_1}
\end{table}

\begin{table}[ht]
\centering
\footnotesize
\setlength{\tabcolsep}{3mm}{ %Change weight of table
\begin{tabular}{lccc}
\toprule
\textit{\textbf{IEMOCAP 4-way}}     
& \multicolumn{1}{c}{Train} & \multicolumn{1}{c}{\,\,\,\,Valid} & Test \\ 
\midrule 
\midrule
\multicolumn{1}{l}{Happy}    & \multicolumn{1}{c}{453}  
& \multicolumn{1}{c}{51}   & \multicolumn{1}{c}{144}  \\ 
\multicolumn{1}{l}{Sad}    & \multicolumn{1}{c}{783}  
& \multicolumn{1}{c}{56}  & \multicolumn{1}{c}{245} \\ 
% \midrule
\multicolumn{1}{l}{Neutral}  & \multicolumn{1}{c}{1,092}
& \multicolumn{1}{c}{232}  & \multicolumn{1}{c}{384}\\ 
\multicolumn{1}{l}{Angry}  & \multicolumn{1}{c}{872} 
& \multicolumn{1}{c}{61}  & \multicolumn{1}{c}{170} \\
\midrule
\multicolumn{1}{l}{Total}  & \multicolumn{1}{c}{3,200} 
& \multicolumn{1}{c}{400}  & \multicolumn{1}{c}{943} \\ 
\bottomrule
\end{tabular}}
\caption{Labels distribution of IEMOCAP 4-way.} 
%IEMOCAP shows the number of utterances/conversations. }
\label{data_2}
\end{table}

\begin{table}[ht]
\centering
\footnotesize
\setlength{\tabcolsep}{3mm}{ %Change weight of table
\begin{tabular}{lccc}
\toprule
\textit{\textbf{IEMOCAP 6-way}}     
& \multicolumn{1}{c}{Train} & \multicolumn{1}{c}{\,\,\,\,Valid} & Test \\ 
\midrule 
\midrule
\multicolumn{1}{l}{Happy}    & \multicolumn{1}{c}{459}  
& \multicolumn{1}{c}{45}   & \multicolumn{1}{c}{144}  \\ 
\multicolumn{1}{l}{Sad}    & \multicolumn{1}{c}{746}  
& \multicolumn{1}{c}{93}  & \multicolumn{1}{c}{245} \\ 
% \midrule
\multicolumn{1}{l}{Neutral}  & \multicolumn{1}{c}{1,161}
& \multicolumn{1}{c}{163}  & \multicolumn{1}{c}{384}\\ 
\multicolumn{1}{l}{Angry}  & \multicolumn{1}{c}{854} 
& \multicolumn{1}{c}{79}  & \multicolumn{1}{c}{170} \\
\multicolumn{1}{l}{Excited}  & \multicolumn{1}{c}{576} 
& \multicolumn{1}{c}{166}  & \multicolumn{1}{c}{299} \\
\multicolumn{1}{l}{Frustrated}  & \multicolumn{1}{c}{1,350} 
& \multicolumn{1}{c}{118}  & \multicolumn{1}{c}{381} \\
\midrule
\multicolumn{1}{l}{Total}  & \multicolumn{1}{c}{5,146} 
& \multicolumn{1}{c}{644}  & \multicolumn{1}{c}{1,623} \\ 
\bottomrule
\end{tabular}}
\caption{Labels distribution of IEMOCAP 6-way.} 
%IEMOCAP shows the number of utterances/conversations. }
\label{data_3}
\end{table}

\begin{table}[!ht]
\centering
\footnotesize
\setlength{\tabcolsep}{4mm}{ %Change weight of table
\begin{tabular}{lccc}
\toprule
\textit{\textbf{MOSEI}}     
& \multicolumn{1}{c}{Train} & \multicolumn{1}{c}{\,\,\,\,Valid} & Test \\ 
\midrule 
\midrule
\multicolumn{1}{l}{Happy}    & \multicolumn{1}{c}{8,735}  
& \multicolumn{1}{c}{1,005}   & \multicolumn{1}{c}{2,505}  \\ 
\multicolumn{1}{l}{Sad}    & \multicolumn{1}{c}{4,269}  
& \multicolumn{1}{c}{520}  & \multicolumn{1}{c}{1,129} \\ 
% \midrule
\multicolumn{1}{l}{Angry}  & \multicolumn{1}{c}{3,526}
& \multicolumn{1}{c}{338}  & \multicolumn{1}{c}{1,071}\\ 
\multicolumn{1}{l}{Surprise}  & \multicolumn{1}{c}{1,642} 
& \multicolumn{1}{c}{203}  & \multicolumn{1}{c}{441} \\ 
\multicolumn{1}{l}{Disgusted}  & \multicolumn{1}{c}{2,955} 
& \multicolumn{1}{c}{281}  & \multicolumn{1}{c}{805} \\ 
\multicolumn{1}{l}{Fear}  & \multicolumn{1}{c}{1,331}
& \multicolumn{1}{c}{176}  & \multicolumn{1}{c}{385} \\ 
\midrule
\multicolumn{1}{l}{Total}  & \multicolumn{1}{c}{22,458}
& \multicolumn{1}{c}{2,523}  & \multicolumn{1}{c}{6,336} \\ 
\bottomrule
\end{tabular}}
\caption{Labels distribution of MOSEI dataset.} 
%IEMOCAP shows the number of utterances/conversations. }
\label{data_4}
\end{table}

\begin{table*}[!t]
\begin{center}
\setlength\tabcolsep{4pt}
\begin{tabular}{lccccc}
\toprule
\multirow{2}{*}{\textbf{Case}} & \multicolumn{3}{c}{\textbf{Input modality}}                 & \multicolumn{2}{c}{\textbf{Target}} 
\\ 
\cmidrule(lr){2-4}\cmidrule(l){5-6}
\multicolumn{1}{c}{}
& Text &Visual &Acoustic & MSA & \,\,MERC  \\
\midrule
\midrule
Case A 
& \multicolumn{1}{c}{\makecell[l]{\footnotesize{Plot to it than that the action scenes were} \\ \footnotesize{\textcolor{blue}{\underline{my favorite parts through it's.}}}}} 
& \multicolumn{1}{c}{\makecell[l]{\underline{\footnotesize{Smiling face}} \\ \underline{\footnotesize{Relaxed wink}}}}
& \multicolumn{1}{c}{\makecell[l]{\,\,\,\,\,\underline{\footnotesize{Stress}} \\ \underline{\footnotesize{Pitch variation}}}}
&+1.666 & \,\,Positive \\
\midrule
Case B 
& \multicolumn{1}{c}{\makecell[l]{\footnotesize{You must promise me that you'll survive,} \\ \footnotesize{\textcolor{blue}{you won't give up.}}}} 
& \multicolumn{1}{c}{\makecell[l]{\underline{\footnotesize{Full of tears }} \\ \underline{\footnotesize{in his eyes}}}}
& \multicolumn{1}{c}{\makecell[l]{\,\,\,\,\footnotesize{The voice is} \\ \underline{\footnotesize{weak and trembling}}}}
&-1.200 & \,\,Negative \\
\bottomrule
\end{tabular}
\end{center}
\caption{Case study on the importance of each modality for MSA and MERC tasks. \textcolor{blue}{Blue} in Text modality marks the contents including the strength of sentiments.
\underline{Underline} marks fragments contributing to the target on MERC.
} 
\label{case_study}
\end{table*}

% \clearpage
\section{Multimodal Sentiment Analysis}
\label{MSA}

We conducted experiments on two publicly available datasets, \textbf{MOSI}~\citep{7742221} and \textbf{MOSEI}~\citep{MorencyCPLZ18}, to investigate the performance of \textsc{Joyful} on the multimodal sentiment analysis (MSA) task.

\noindent \textcolor{violet}{\adfhalfrightarrowhead} \textcolor{violet}{\textbf{Datasets:}}
MOSI contains 2,199 utterance video segments, and each segment is manually annotated with a sentiment score ranging from -3 to +3 to indicate the sentiment polarity and relative sentiment strength of the segment.
MOSEI contains 22,856 movie review clips from the YouTube website. Each clip is annotated with a sentiment score and an emotion label.  
And the exact number of samples for training/validation/test are 1,284/229/686 for MOSI and 16,326/1,871/4,659 for MOSEI.

\noindent  \textcolor{violet}{\adfhalfrightarrowhead}  \textcolor{violet}{\textbf{Metrics:}} 
Following previous studies~\citep{HanCG0MP21,YuXYW21},
we utilized evaluation metrics:
mean absolute error (MAE) measures the absolute error between predicted and true values.
Person correlation (Corr) measures the degree of prediction skew. 
Seven-class classification accuracy (ACC-7) indicates the proportion of predictions that correctly fall into the same interval of seven intervals between -3 and +3 as the corresponding truths.
And binary classification accuracy (ACC-2) was computed for non-negative/negative classification results.

\noindent \textcolor{violet}{\adfhalfrightarrowhead}  \textcolor{violet}{\textbf{Baselines:}} 
We compared \textsc{Joyful} with three types of advanced multimodal fusion frameworks for the MSA task as follows, including current SOTA baselines MMIM~\citep{HanCP21} and BBFN~\citep{HanCG0MP21}: (1) Early multimodal fusion methods, which combine the different modalities before they are processed by any neural network models. 
We utilized Multimodal Factorization Model (\textbf{MFM})~\citep{MFM}, and Multimodal Adaptation Gate BERT (\textbf{MAG-BERT})~\citep{RahmanHLZMMH20} as baselines.
(2) Late multimodal fusion methods, which combine the different modalities before the final decision or prediction layer. 
We utilized multimodal Transformer (\textbf{MuIT})~\citep{Mult}, and modal-temporal attention graph (\textbf{MTAG})~\citep{YangWYZRZPM21} as baselines.
(3) Hybrid multimodal fusion methods combine early and late multimodal fusion mechanisms to capture the consistency and the difference between different modalities simultaneously. 
We utilized modality-invariant and modality-specific representations for MSA (\textbf{MISA})~\citep{MISA_paper}, Self-Supervised multi-task learning for MSA (\textbf{Self-MM})~\citep{YuXYW21}, Bi-Bimodal Fusion Network (\textbf{BBFN})~\citep{HanCG0MP21}, and MultiModal InfoMax (\textbf{MMIM})~\citep{HanCP21} as baselines.

\noindent \textcolor{violet}{\adfhalfrightarrowhead} \textcolor{violet}{\textbf{Implementation Details:}} 
The results of proposed \textsc{Joyful} were averaged over ten runs using random seeds.
We keep all hyper-parameters and implementations the same as in the MERC task reported in Sections 4.1 and 4.2.
To make \textsc{Joyful} fit in the MSA task, we replace the current cross-entropy loss $\mathcal{L}_{ce}$ in Eq. (15) by mean absolute error loss $\mathcal{L}_{mae}$ as follows:
\begin{equation}
\footnotesize
    \mathcal{L}_{mae} = \frac{1}{m} \sum_{i=1}^{m} |\hat{y}_{i} - y_{i}|,
\end{equation}
where $\hat{y}_{i}$ is the predicted value for the $i$-th sample, $y_{i}$ is the truth label for the $i$-th label, $m$ is the total number of samples, and $|\cdot|$ is the $L_{1}$ norm. We denote this model as \textsc{Joyful+MAE}.

\begin{table}[!ht]
\scriptsize
\centering
\setlength\tabcolsep{0.05pt}
\begin{tabular}{lcccccccccc}
\toprule
\multirow{2}{*}{\textit{\textbf{Method}}} & \multicolumn{4}{c}{\textbf{MOSI}}         &\,\,       & \multicolumn{4}{c}{\textbf{MOSEI}} \\ 
\cmidrule(lr){2-5}\cmidrule(l){7-10}
\multicolumn{1}{c}{}
& MAE $\downarrow$ & Corr $\uparrow$ & Acc-7 $\uparrow$ & Acc-2 $\uparrow$ &\,\, & MAE $\downarrow$ & Corr $\uparrow$ & Acc-7 $\uparrow$ & Acc-2 $\uparrow$ \\
\midrule
\midrule
MFM\,\,
& 0.877 & 0.706 & 35.4 & 81.7 &\,\,
& 0.568 & 0.717 & 51.3 & 84.4 \\
MAG-BERT\,\,
& 0.731 & 0.789 & \ding{55} & 84.3 &\,\,
& 0.539 & 0.753 & \ding{55} & 85.2 \\
MulT\,\, 
& 0.861 & 0.711 & \ding{55} & 84.1 &\,\,
& 0.580 & 0.703 & \ding{55} & 82.5 \\
MTAG\,\, 
&0.866 &0.722 &0.389 &82.3 &\,\,
&\ding{55} &\ding{55} &\ding{55} &\ding{55}\\
MISA\,\,
& 0.804 & 0.764 & \ding{55} & 82.10 &\,\,
& 0.568 & 0.724 & \ding{55} & 84.2 \\
Self-MM\,\, 
& 0.713 & 0.789 & \ding{55} & 85.98 &\,\,
& 0.530 & 0.765 & \ding{55} & 85.17 \\
\specialrule{0em}{4pt}{0pt} 
\arrayrulecolor{blue}
\midrule
\specialrule{0em}{0pt}{4pt}
\arrayrulecolor{black}
% \midrule
BBFN\,\,
& 0.776 & 0.755 & 45.00 & 84.30 &\,\,
& 0.529 & 0.767 & \textbf{54.80} & \textbf{86.20} \\
MMIM\,\,
& \textbf{0.700} & \textbf{0.800} & \textbf{46.65} & \textbf{86.06} &\,\,
& \textbf{0.526} & \textbf{0.772} & 54.24 & 85.97 \\
\midrule
\textsc{Joyful+MAE} 
& 0.711 & 0.792 & 45.58 & 85.87 &\,\,
& 0.529 & 0.768 & 53.94 & 85.68 \\
\bottomrule
\end{tabular}
%\caption{Multimodal sentiment analysis performance on MOSI and MOSEI datasets, where bold means SOTA performance.} 
\caption{Experimental results on the MOSI and MOSEI datasets. 
%\ding{55} represents that they have not reported the results. 
\ding{55} indicates unreported results. 
\textbf{Bold} indicates the least MAE, highest Corr, Acc-7, and Acc-2 scores for each dataset.} 
\label{MSA_performance}
\end{table}

Experimental results on the MOSI and MOSEI datasets are listed in Table~\ref{MSA_performance}. 
Although the proposed \textsc{Joyful} could outperform most of the baselines (above the blue line), it performs worse than current SOTA models:  BBFN and MMIM (below the blue line). 
We conjecture the main reasons are:
when determining the strength of sentiments, compared with visual and acoustic modalities that may contain much noise data, text modality is more important for prediction~\citep{HanCG0MP21}. Table~\ref{case_study} lists such examples, where textual modality is more indicative than other modalities for the MSA task.
Because the two baselines: BBFN~\citep{HanCG0MP21} and MMIN~\citep{HanCP21}, pay more attention to the text modality than visual and acoustic modalities during multimodal feature fusion, they may achieve low MAE, high Corr, Acc-2, and Acc-7. 
Specifically, BBFN~\citep{HanCG0MP21} proposed a Bi-bimodal fusion network to enhance the text modality's importance by only considered text-visual and text-acoustic interaction for features fusion. 
Conversely, considering the three modalities are all important for the MERC task as presented in Table~\ref{case_study}, we designed \textsc{Joyful} to utilize the concatenation of the three modalities representations for prediction. 
Similar to our proposal, MISA and MAG-BERT considered the three modalities equally important during feature fusion but performed worse than SOTA baselines on the MSA task.
In our consideration, because of such attention to modalities, \textsc{Joyful} outperformed SOTA baselines on the MERC task but underperformed SOTA baselines on the MSA task.

\end{document}